\documentclass{article}

\usepackage{PRIMEarxiv}
\usepackage[utf8]{inputenc} 
\usepackage{csvsimple}
\usepackage{booktabs}
\usepackage{siunitx}
\usepackage{longtable}
\usepackage{graphicx}
\usepackage{subcaption}
\usepackage{float}

\usepackage[hidelinks]{hyperref}
\usepackage{xr}  

\usepackage{soul}
\usepackage{xcolor}

\usepackage{booktabs}
\usepackage{natbib}%
 \usepackage{graphics}
 \usepackage{tabularx}
\usepackage{makecell}
\usepackage[export]{adjustbox}
\usepackage{multirow}
\usepackage{enumitem}
\setlist{labelindent=12pt}
\usepackage{subcaption}
\usepackage{algorithm} 
\usepackage{algpseudocode}
\usepackage[labelfont=bf,textfont={bf}]{caption}
\usepackage{xcolor}
\usepackage{nccmath, mathtools}
\usepackage{svg}
\usepackage{amssymb} 
\usepackage{etoolbox}
\apptocmd\normalsize{%
 \abovedisplayskip=12pt
 \abovedisplayshortskip=0pt 
 \belowdisplayskip=12pt 
 \belowdisplayshortskip=7pt
}{}{}
\usepackage{mathtools}
\usepackage{physics}
\hypersetup{
    colorlinks=true,                
    breaklinks=true,                
    urlcolor= cyan,                
    linkcolor= red,                     
    bookmarksopen=false,
    filecolor=black,
    citecolor=blue,
    linkbordercolor=pink
}

\renewcommand{\mathbf}{\boldsymbol}

\newcount\Comments  

\definecolor{darkgreen}{rgb}{0,0.5,0}
\definecolor{purple}{rgb}{1,0,1}
\definecolor{darkblue}{rgb}{0.6,0.4,0.8}

\newcommand{\kibitz}[2]{\ifnum\Comments=0\textcolor{#1}{#2}\fi}

\title{A phase-aware AI car-following model for electric vehicles with adaptive cruise control: Development and validation using real-world data

\thanks{\textit{\underline{Citation}}: 
\textbf{Liu et al. A phase-aware AI car-following model for electric vehicles with adaptive cruise control: Development and validation using real-world data.}} 
}

\author{
  Yuhui Liu \\
  Department of Civil Engineering \\
  Saint Louis University  \\
  \AND
  Shian Wang \\
 Department of Civil, Environmental, and Architectural Engineering\\
  University of Kansas\\  
  \AND
  Ansel Panicker \\
  Department of Computer Science\\
  Saint Louis University   \\
    \AND
  Kate Embry \\
  Department of Civil Engineering \\
  Saint Louis University  \\
  \AND
  Ayana Asanova\\
  Department of Civil Engineering \\
  Saint Louis University  \\
  \AND
    Tianyi Li \\
  Department of Civil Engineering \\
  Saint Louis University \\
  \texttt{tianyili.ai@gmail.com}}

\begin{document}
\maketitle

\begin{abstract}
Internal combustion engine (ICE) vehicles and electric vehicles (EVs) exhibit distinct vehicle dynamics. EVs provide rapid acceleration, with electric motors producing peak power across a wider speed range, and achieve swift deceleration through regenerative braking. While existing microscopic models effectively capture the driving behavior of ICE vehicles, a modeling framework that accurately describes the unique car-following dynamics of EVs is lacking. Developing such a model is essential given the increasing presence of EVs in traffic, yet creating an easy-to-use and accurate analytical model remains challenging.

To address these gaps, this study develops and validates a Phase-Aware AI (PAAI) car-following model specifically for EVs. The proposed model enhances traditional physics-based frameworks with an AI component that recognizes and adapts to different driving phases, such as rapid acceleration and regenerative braking. Using real-world trajectory data from vehicles equipped with adaptive cruise control (ACC), we conduct comprehensive simulations to validate the model's performance. The numerical results demonstrate that the PAAI model significantly improves prediction accuracy over traditional car-following models, providing an effective tool for accurately representing EV behavior in traffic simulations.

\end{abstract}


\section{Introduction}{\label{sec:intr}}

The global adoption of electric vehicles (EVs) has accelerated over the past decade, driven by environmental concerns, government subsidies, and technological advancements~\citep{iea2023,bloom2023}. Compared to traditional internal combustion engine (ICE) vehicles, EVs offer significant advantages, such as higher energy efficiency and zero tailpipe emissions~\citep{hawkins2012environmental,acedo2025impacts}. From a performance standpoint, EVs also exhibit distinct vehicle dynamics: they provide rapid acceleration, as electric motors generate peak power over a wide speed range, and achieve swift deceleration through regenerative braking~\citep{chan2007state}. As the market penetration of EVs continues to increase, these unique characteristics hold important implications for traffic flow and road safety~\citep{dnv2024,ahmed2024systematic}.

Despite the growing presence of EVs, current microscopic traffic models have been predominantly developed to capture the driving behavior of ICE vehicles~\citep{gazis1961nonlinear, stern2018dissipation,li2024car,li2021classification,zhang2024car}. While these models effectively describe traditional powertrain, they fail to fully account for the unique acceleration and deceleration characteristics of EVs~\citep{treiber2000congested}. The absence of a dedicated modeling framework for the car-following dynamics of EVs presents a critical gap in the literature, especially given how essential microscopic models are for understanding and managing traffic flow~\citep{bando1995dynamical}.

Developing a robust yet easy-to-use car-following model for EVs is therefore both timely and challenging. It requires careful treatment of an electric powertrain's instantaneous torque delivery and regenerative braking mechanisms, which differ fundamentally from the internal combustion process. Furthermore, the increasing prevalence of advanced driver-assistance systems (ADAS)—particularly adaptive cruise control (ACC)—presents an opportunity to refine our understanding of vehicle-following behavior~\citep{talebpour2016influence,kesting2008adaptive}. ACC-equipped vehicles generate high-fidelity trajectory data that capture real-world driving conditions, enabling the development of more accurate and predictive microscopic traffic models.

A key contribution of this study is our detailed comparative analysis of the car-following behaviors of EVs and ICE vehicles, based on the insights from related work summarized in Table~\ref{tab:literature_classification}. We leverage extensive real-world EV ACC trajectory data to characterize how EVs accelerate and decelerate when following lead vehicles~\citep{yang2024microsimacc} and to compare these dynamics directly with the established behaviors of ICE vehicles~\citep{li2025racer, li2024detecting}.

Building on the distinct behavioral differences identified in this analysis, we introduce our primary contribution: the proposed Phase-Aware AI (PAAI) model. This model improves upon traditional car-following frameworks by specifically capturing EV dynamics, such as rapid torque response and regenerative braking. These enhancements enable the model to reveal how EV behaviors influence congestion patterns and roadway efficiency, providing quantitative insights into their macroscopic impacts and offering guidance for future transport policies and infrastructure planning.

\begin{table}[t!]
\captionsetup{justification=centering}
\caption{Summary of literature by research domain.}
\centering
\small
\setlength{\tabcolsep}{4pt}
\begin{tabular}{@{}p{4.8cm}cp{4.4cm}p{4.0cm}@{}}
\toprule
\textbf{Study} & \textbf{Year} & \textbf{Key Findings} & \textbf{Performance/Impact} \\
\midrule
\multicolumn{4}{l}{\textit{\textbf{EV Adoption and Benefits}}} \\
\midrule
~\cite{iea2023} & 2023 & Global EV adoption trends & Shapes environmental policy \\
~\cite{bloom2023} & 2023 & EV market growth & Guides industry investment \\
~\cite{hawkins2012environmental} & 2012 & EV efficiency, no emissions & Supports sustainability claims \\
~\cite{acedo2025impacts} & 2025 & EV environmental benefits & Informs future EV policy \\
\addlinespace
\midrule
\multicolumn{4}{l}{\textit{\textbf{EV Dynamics}}} \\
\midrule
~\cite{chan2007state} & 2007 & EV acceleration, braking & Informs EV performance models \\
~\cite{dnv2024} & 2024 & EV traffic flow impact & Enhances safety analysis \\
~\cite{ahmed2024systematic} & 2024 & EV traffic performance & Guides traffic management \\
\addlinespace
\midrule
\multicolumn{4}{l}{\textit{\textbf{Microscopic Traffic Modeling}}} \\
\midrule
~\cite{gazis1961nonlinear} & 1961 & Early ICE car-following & Foundational traffic theory \\
~\cite{stern2018dissipation} & 2018 & ICE traffic flow models & Improves congestion analysis \\
~\cite{li2021classification,li2024car} & 2021, 2024 & ICE model limits for EVs & Highlights EV modeling gap \\
~\cite{zhang2024car} & 2024 & Surveys ICE car-following & Identifies research gaps \\
~\cite{treiber2000congested} & 2000 & ICE model EV limits & Supports need for EV models \\
~\cite{bando1995dynamical} & 1995 & Microscopic traffic control & Enhances flow prediction \\
\addlinespace
\midrule
\multicolumn{4}{l}{\textit{\textbf{AI and ACC in Traffic Modeling}}} \\
\midrule
~\cite{talebpour2016influence} & 2016 & ACC in car-following & Improves model accuracy \\
~\cite{kesting2008adaptive} & 2008 & ACC data for traffic & Enables data-driven models \\
~\cite{yang2024microsimacc} & 2024 & ACC data for EV models & Improves EV behavior prediction \\
~\cite{li2025racer,li2024detecting} & 2023, 2024 & AI for EV-ICE comparison & Enhances mixed traffic models \\
\bottomrule
\end{tabular}
\label{tab:literature_classification}
\end{table}

In what follows, Section~\ref{sec:review} reviews the underlying differences in EV and ICE powertrain dynamics and surveys existing car-following models. Section~\ref{sec:data} details the data used for both ACC-equipped ICE vehicles and EVs. Section~\ref{sec:methodology} presents the methodology for developing our AI-based model. Section~\ref{sec:simulations} describes the simulation experiments and results, highlighting how EV deployment at various penetration levels influences traffic performance. Finally, Section~\ref{sec:conclusion} summarizes our key findings and discusses potential directions for future research on EV-centric traffic modeling.

\section{Related Work}\label{sec:review}

\subsection{Evolution of Car-Following (CF) Models}

Car-following (CF) models serve as foundational tools for microscopic traffic flow modeling. One of the earliest nonlinear models introduced the ``Follow-the-Leader'' concept, describing driver responses to leading vehicles~\citep{gazis1961nonlinear}. Subsequently, a dynamical model was proposed to simulate mechanisms underlying traffic congestion~\citep{bando1995dynamical}. Comprehensive reviews have documented the evolution of CF models from empirical observations to theoretical frameworks~\citep{brackstone1999car}. Systematic analyses have further explored the strengths and limitations of classical CF models in real-world applications~\citep{yin2022systematic}.

Over decades, reviews and surveys have chronicled this evolution. For instance,~\cite{li2012microscopic} provide a comprehensive survey of classic CF models, analyzing their development and stability. They examine the full velocity difference (FVD) model by conducting local, asymptotic, and nonlinear stability analyses to characterize traffic flow characteristics. The review highlights extensions such as optimal-velocity and velocity-difference models, which enhance realism and stability. Various CF models, including fuzzy-logic, collision avoidance, and cellular automation models, are also discussed, emphasizing the importance of incorporating multiple headways, velocity differences, and acceleration differences for future research to improve traffic flow stability.

Mathematical modeling plays a crucial role in understanding single-lane CF behavior.~\cite{rothery1992car} focuses on the development and analysis of various CF models, examining their stability and the link between microscopic driver behavior and macroscopic traffic flow. The study includes experimental validation by comparing real-world data with theoretical assumptions and explores automated CF concepts, bridging individual driver actions with collective traffic dynamics.

Beyond one-dimensional freeway models, researchers have developed two-dimensional (2D) CF models to capture lateral interactions.~\cite{delpiano2020two} introduce a 2D CF formulation based on the social-force paradigm, explicitly incorporating collision avoidance and lane-changing behavior. This accounts for lateral friction in high-occupancy vehicle (HOV) lanes, relaxation at merge bottlenecks, and lane-change accidents. Simulations demonstrate that traditional one-lane models, which discretize lateral motion, overlook key phenomena, whereas the 2D approach captures indirect effects of collision avoidance, underscoring its role in multi-lane dynamics.

More recently, CF modeling has become increasingly multidisciplinary and data-driven.~\cite{zhang2024car} survey CF algorithms, ranging from physics-based (kinematic and psychological) to AI-driven and generative models. Models are classified by principles, with emphasis on integrating human factors, vehicle mechanics, and machine intelligence. For example,~\cite{shang2022novel,shang2024two} propose asymmetric CF models for ACC vehicles, while~\cite{han2022modeling} examine the interplay of driver behavior, vehicle properties, and environmental factors. This multidisciplinary trend is essential as vehicles become increasingly automated and connected.

\subsection{Modeling Challenges: ACC and EV Dynamics}

Recent studies emphasize the impact of ACC vehicles on traffic and the need for precise model calibration.~\cite{de2021calibrating} argue that adopting a CF model alone is insufficient; calibration to real data is critical for understanding ACC effects. They employ multi-objective calibration for three CF models, minimizing trade-offs between speed and spacing errors using trajectory data from seven commercial ACC vehicles. Results show that calibration choices significantly affect predictions, providing parameter sets for future simulations and ACC penetration analyses.~\cite{wang2023car} offer a systematic review of CF models for human-driven vehicles (HDVs) and autonomous vehicles (AVs). Models are categorized by assumptions (theory-based vs. AI-driven) and application domains, with discussions on accuracy, continuity, stability, and challenges in complex environments. The review identifies gaps in calibration and stability analysis for realistic traffic, offering suggestions for future research in mixed HDV-AV flows.

The adoption of EVs introduces unique challenges for CF modeling due to their distinct acceleration and deceleration characteristics, such as near-instantaneous torque and regenerative braking, compared to ICE vehicles. These differences must be accounted for in traffic simulations~\citep{he2020introducing}. While a new model has been proposed to capture the CF behavior of EVs using experimental data~\citep{zare2024electric}, it relies on assumptions regarding transitions between acceleration and deceleration modes that may be difficult to validate.

Research on EV technology advancements provides overviews of battery evolution from lead-acid to lithium-ion, charging methods, and vehicle-to-everything (V2X) technologies, offering insights for future development~\citep{sun2019technology}. A recent review examines EV market penetration, battery technologies, charging infrastructure (including standards and energy management), and future directions for battery enhancements, charging processes, smart city integration, and sustainability~\citep{sanguesa2021review}. Earlier work from 2009 surveys key components like energy storage, electric motors, braking systems, charging infrastructure, and prototypes, providing a state-of-the-art snapshot~\citep{cheng2009recent}. These advancements in battery technology enable EVs to achieve rapid acceleration and efficient regenerative braking, which introduce unique CF dynamics compared to ICE vehicles. These battery-driven performance characteristics pose significant modeling challenges for ACC systems, traditional CF models must now account for the nonlinear acceleration profiles and asymmetric braking responses unique to EVs, necessitating new approaches for accurate traffic simulation in mixed-vehicle environments.

Recent field experiments on EVs with ACC show they can maintain shorter headways during speed changes, improving traffic flow and stability, though results vary by manufacturer~\citep{lapardhaja2023unlocking}. The Electric Vehicle Model (EVM) outperforms traditional CF models in predicting ACC-equipped EV behavior, reducing traffic oscillations and enhancing stability in simulations~\citep{zare2024electric}.

\subsection{AI in Transportation and Car-Following}

AI is essential in transportation research for addressing complex, dynamic challenges that traditional methods cannot effectively handle. Conventional analytical models struggle to capture nonlinear vehicle interactions, diverse driver behaviors, and real-time traffic changes, particularly for EVs with rapid acceleration and regenerative braking. AI methods, using machine learning and deep learning, can effectively process large datasets to model these complexities, providing adaptive, predictive, and reliable traffic solutions. In traffic management, AI is widely used to analyze vast datasets and predict traffic flow in real-time, forming the technological backbone of modern navigation and control systems~\citep{gilmore1993ai, gebre2024ai}. In microscopic simulation, machine learning refines models by predicting specific driver behaviors, such as lane changes~\citep{sayed2023artificial}. AI is also integral to optimizing EV performance. For example, deep learning is applied in battery management systems for state-of-charge estimation and fault detection~\citep{badran2024employment}, while other machine learning approaches predict and manage battery thermal performance~\citep{amer2024electric}. Beyond vehicles, AI enhances grid management, optimizes charging strategies, and enables key ADAS safety features through V2X communication~\citep{amer2024electric}.

Within CF modeling specifically, AI has been used to create more adaptive and human-like driving behaviors. Deep reinforcement learning, for example, has been applied to develop models that improve safety and operational efficiency in AVs~\citep{feng2023dense,masmoudi2021reinforcement,hart2021formulation}. A particularly promising direction is the fusion of physics-based models with data-driven AI. Physics-Informed Neural Networks (PINNs) and frameworks like the Physics-Enhanced Residual Learning (PERL) model combine the interpretability of traditional models with the adaptability of neural networks. These hybrid approaches learn to correct the errors of a baseline physics model, resulting in highly accurate and data-efficient solutions~\citep{long2024traffic,li2024customizable,li2025racer}. This trend of integrating AI with established traffic theory provides a robust foundation for developing the next generation of CF models capable of capturing the unique dynamics of EVs in mixed traffic with ICE vehicles.

\section{Data Sources}\label{sec:data}

This study employs two high-fidelity datasets to develop and validate an AI-based car-following model: one comprising trajectory data from ICE vehicles, and the other from EVs, both equipped with ACC. These datasets support a robust analysis of car-following dynamics across vehicle types, enabling direct comparisons of their responses in dynamic traffic scenarios. Below, we describe the two datasets, including their sources, experimental protocols, and key characteristics, followed by a comparison of features most relevant to our modeling objectives.

\subsection{ICE Vehicle Data}

\begin{table}[t!]
\captionsetup{justification=centering}
\vspace{1em}
\caption{Summary of tested vehicles.}
\begin{center}
\begin{tabular}{ccccc}
\toprule
Vehicle & Make & Style & Engine & Min. ACC Speed (m/s) \\
\midrule
A & 1 & Full-size sedan & Combustion & 11.18 \\
B & 1 & Compact sedan & Combustion & 11.18 \\
C & 1 & Compact hatchback & Hybrid & 11.18 \\
D & 1 & Compact SUV & Combustion & 11.18 \\
E & 2 & Compact SUV & Combustion & 0.00 \\
F & 2 & Mid-size SUV & Combustion & 0.00 \\
G & 2 & Full-size SUV & Combustion & 0.00 \\
\bottomrule
\end{tabular}
\end{center}
\label{tab:vehicle_summary}
\end{table}

\begin{figure}[t!]
    \centering
    \begin{subfigure}{0.24\textwidth}
        \centering
        \includegraphics[width=\textwidth]{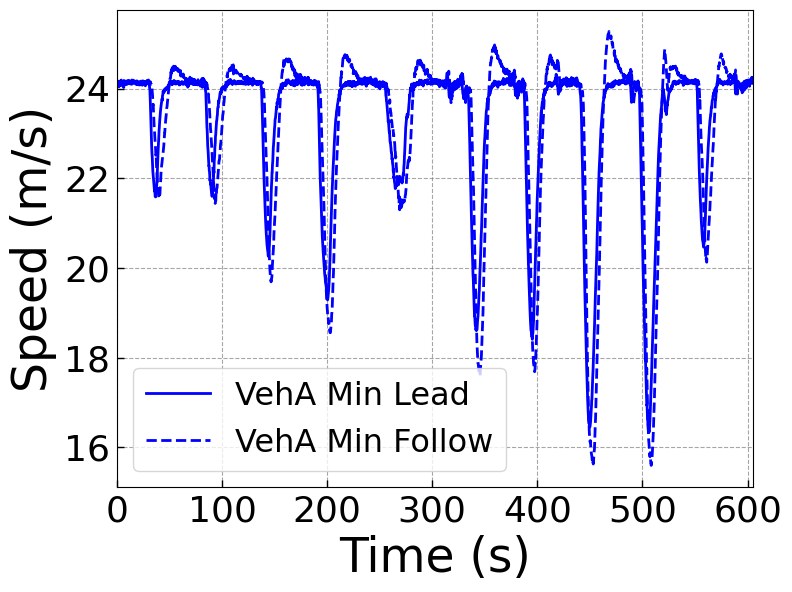}
        \caption{Speed (Veh A Min)}
        \label{fig:ice_speed_a_min}
    \end{subfigure}
    \hfill
    \begin{subfigure}{0.24\textwidth}
        \centering
        \includegraphics[width=\textwidth]{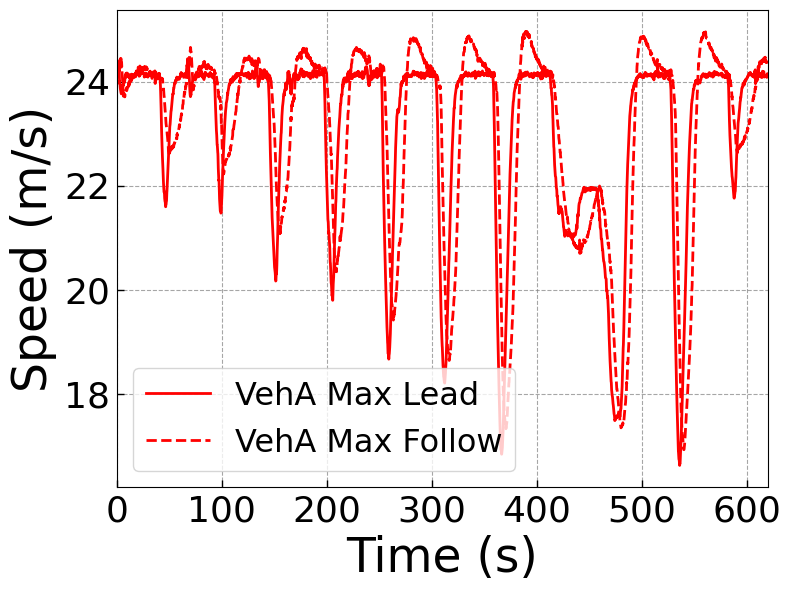}
        \caption{Speed (Veh A Max)}
        \label{fig:ice_speed_a_max}
    \end{subfigure}
    \hfill
    \begin{subfigure}{0.24\textwidth}
        \centering
        \includegraphics[width=\textwidth]{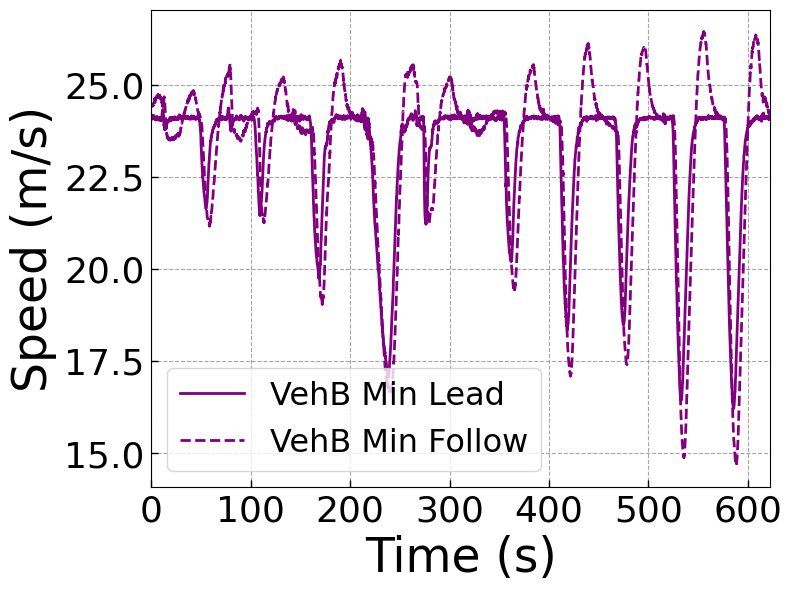}
        \caption{Speed (Veh B Min)}
        \label{fig:ice_speed_b_min}
    \end{subfigure}
    \hfill
    \begin{subfigure}{0.24\textwidth}
        \centering
        \includegraphics[width=\textwidth]{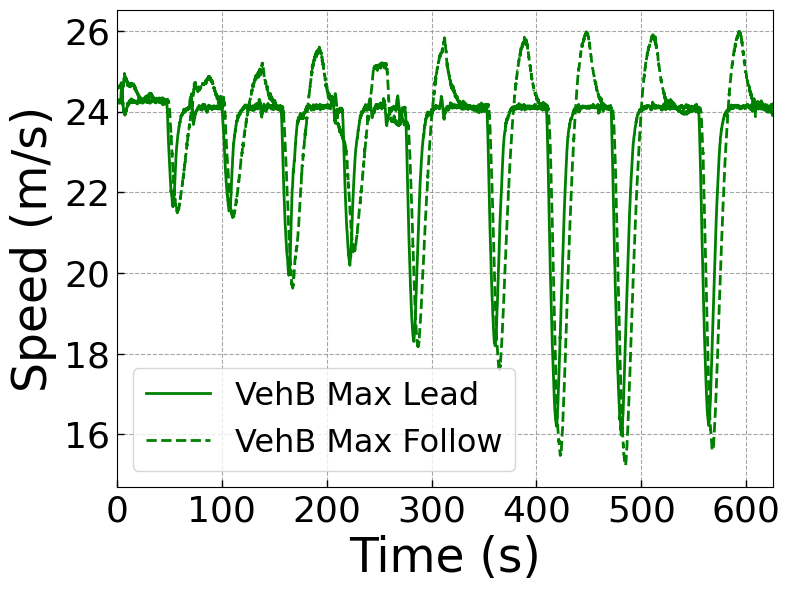}
        \caption{Speed (Veh B Max)}
        \label{fig:ice_speed_b_max}
    \end{subfigure}
    
    \vspace{1em}
    \begin{subfigure}{0.24\textwidth}
        \centering
        \includegraphics[width=\textwidth]{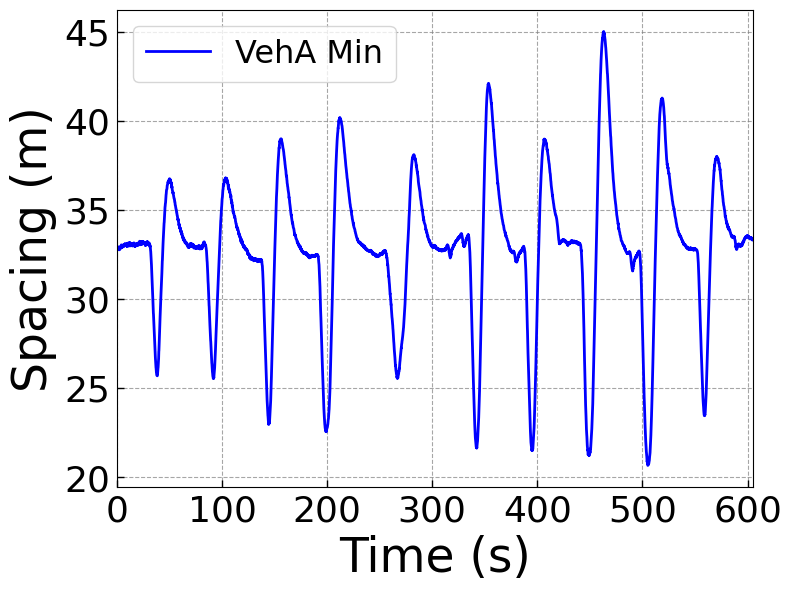}
        \caption{Spacing (Veh A Min)}
        \label{fig:ice_spacing_a_min}
    \end{subfigure}
    \hfill
    \begin{subfigure}{0.24\textwidth}
        \centering
        \includegraphics[width=\textwidth]{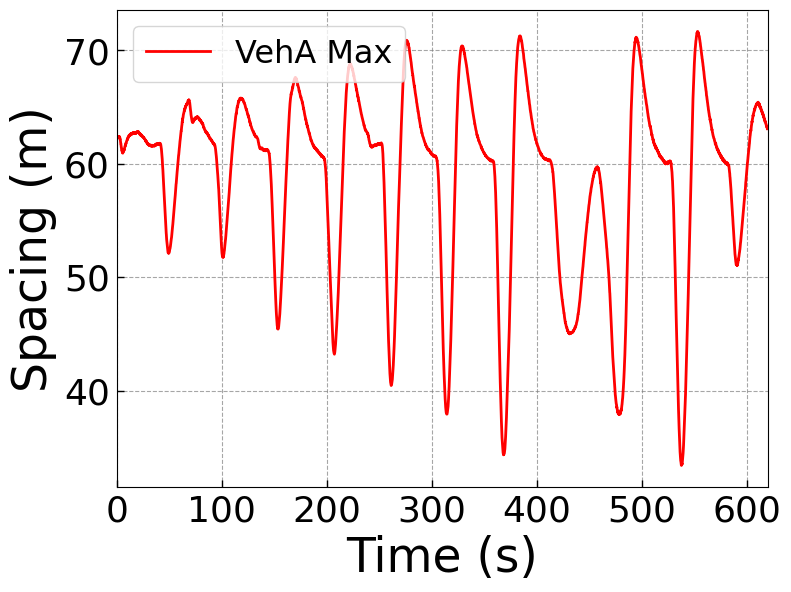}
        \caption{Spacing (Veh A Max)}
        \label{fig:ice_spacing_a_max}
    \end{subfigure}
    \hfill
    \begin{subfigure}{0.24\textwidth}
        \centering
        \includegraphics[width=\textwidth]{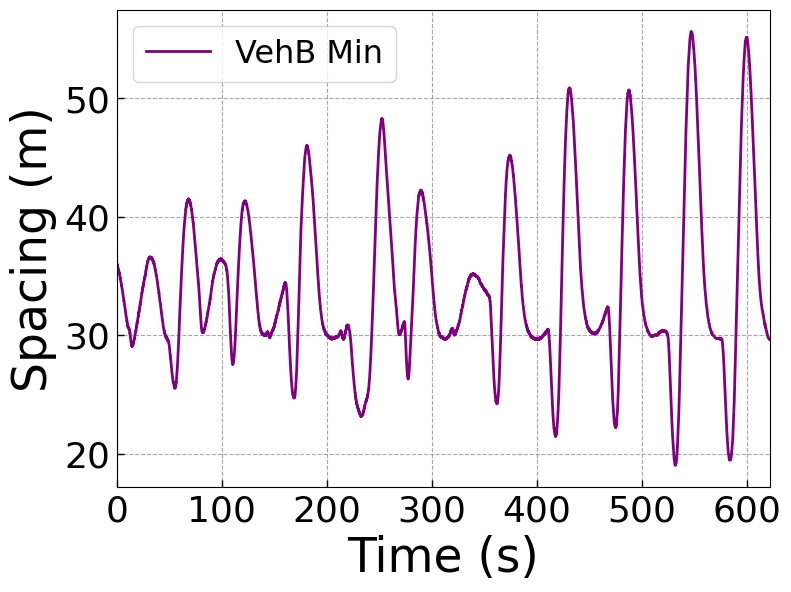}
        \caption{Spacing (Veh B Min)}
        \label{fig:ice_spacing_b_min}
    \end{subfigure}
    \hfill
    \begin{subfigure}{0.24\textwidth}
        \centering
        \includegraphics[width=\textwidth]{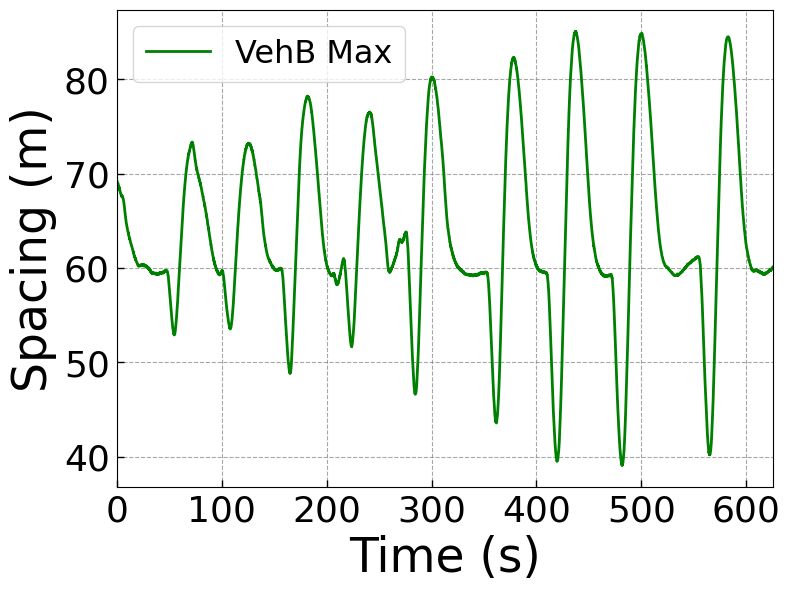}
        \caption{Spacing (Veh B Max)}
        \label{fig:ice_spacing_b_max}
    \end{subfigure}
    \caption{Speed and spacing profiles for ICE vehicles A and B under the `speed dips' condition with minimum and maximum ACC settings. The first row shows speed profiles for Vehicle A (min/max) and Vehicle B (min/max) from left to right; the second row represents spacing profiles for Vehicle A (min/max) and Vehicle B (min/max) from left to right.}
    \label{fig:ice_speed_spacing_dataset}
\end{figure}

The ICE vehicle dataset is derived from two-vehicle CF experiments with 2018 model year vehicles, as reported in~\citep{gunter2020commercially}. It includes trajectory data from seven ACC-equipped vehicles (Vehicles A--G), with six traditional ICE vehicles (A, B, D, E, F, G) and one hybrid (C). Table~\ref{tab:vehicle_summary} summarizes the test vehicles~\citep{gunter2020commercially}. This study utilizes data from ICE Vehicles A and B under minimum (approximately 32 m) and maximum (approximately 60 m) ACC space-gap settings in the `speed dips' condition (``vehA dip min/max'' and ``vehB dip min/max''), capturing transient CF behavior for comparison with EV dynamics. Vehicles A and B were selected, excluding larger SUVs (D, E, F, G) for consistency. This ICE dataset, combined with the EV dataset, enables a comparative analysis to observe differences in CF behavior between ICE vehicles and EVs.

The dataset, chosen for its accurate capture of ACC-equipped ICE dynamics, uses the `speed dips' condition to mirror EV experiment scenarios, involving constant speed, deceleration, and re-acceleration in a CF context. The `speed dips' condition involves a lead vehicle executing speed fluctuations to capture transient CF behavior, selected to assess dynamic responses and align with EV dataset conditions for consistent comparison~\citep{gunter2020commercially}. Its public availability and multiple vehicles support reproducible mixed traffic analysis.

Trajectory data were collected at 10~Hz using uBlox GPS receivers, with position accuracy of 0.24~m and speed accuracy of 0.002~m/s~\citep{gunter2020commercially}. Experiments featured a full-size sedan lead vehicle with a programmed speed profile, followed by an ACC-equipped ICE vehicle in a single-lane platoon on a flat road. The `speed dips' tests captured dynamic responses under maximum ACC settings, selected to align with EV dataset conditions~\citep{lapardhaja2023unlocking}. The ICE subset matches EV data to minimize experimental variations, ensuring a comparable analysis. Speed and spacing profiles for Vehicles A and B are shown in Figure~\ref{fig:ice_speed_spacing_dataset}.

\subsection{EV Data}

This study utilizes a real-world trajectory dataset from recent CF experiments involving an EV equipped with ACC, as reported in~\citep{lapardhaja2023unlocking}. The dataset, collected using a 2022 Hyundai IONIQ 5 as the EV-ACC test vehicle and a 2021 Toyota Camry as the ICE lead vehicle, forms the foundation for our analysis and model development. The dataset was selected for its high-fidelity capture of EV-ACC dynamics, including rapid acceleration and regenerative braking, with varying ACC settings. Its 25~Hz temporal resolution and diverse traffic scenarios support accurate modeling of complex EV interactions in mixed traffic. Experiments involved the EV-ACC following the ICE lead vehicle at a free-flow speed of 24.6~m/s (55~mph), representing typical highway conditions, across four space-gap settings: short (approximately 25~m), medium (approximately 35~m), long (approximately 45~m), and extra-long (approximately 55~m), as defined by initial headway distances.

\begin{figure}[t!]
    \centering
    \begin{subfigure}{0.24\textwidth} 
        \centering
        \includegraphics[width=\textwidth]{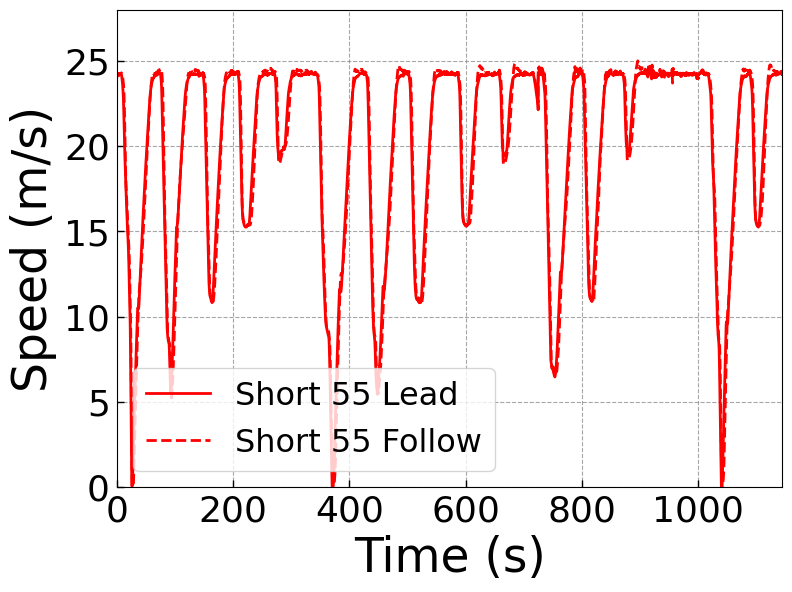}
        \caption{Speed (Short)}
        \label{fig:speed_time_short_55}
    \end{subfigure} 
    \hfill
    \begin{subfigure}{0.24\textwidth}
        \centering
        \includegraphics[width=\textwidth]{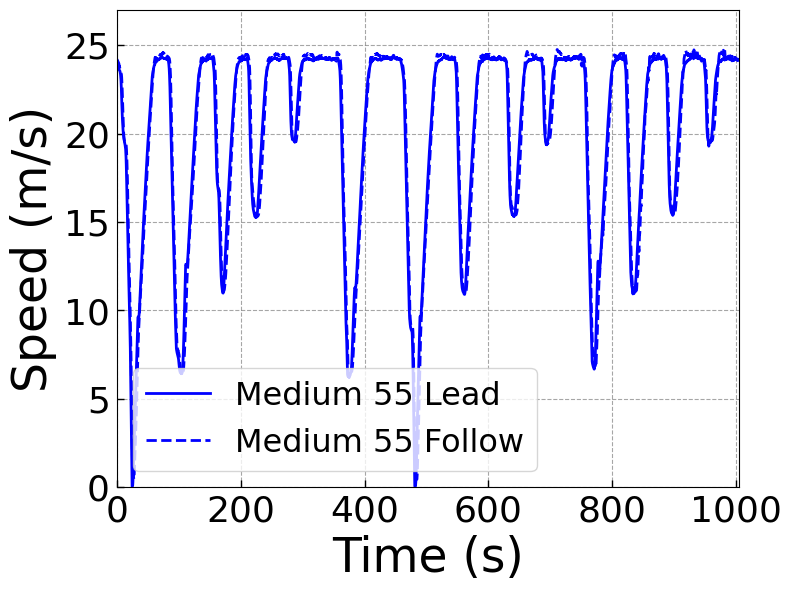}
        \caption{Speed (Medium)}
        \label{fig:speed_time_medium_55}
    \end{subfigure}
    \hfill
    \begin{subfigure}{0.24\textwidth}
        \centering
        \includegraphics[width=\textwidth]{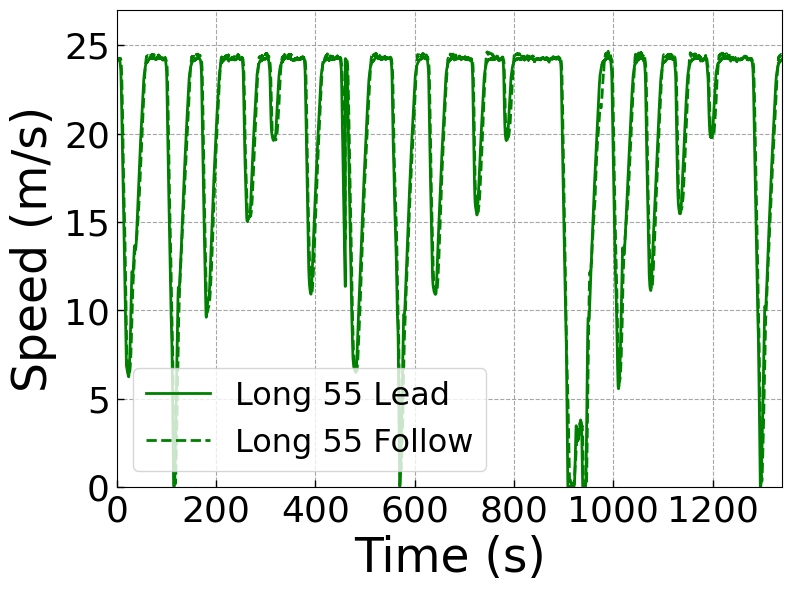}
        \caption{Speed (Long)}
        \label{fig:speed_time_long_55}
    \end{subfigure}
    \hfill
    \begin{subfigure}{0.24\textwidth}
        \centering
        \includegraphics[width=\textwidth]{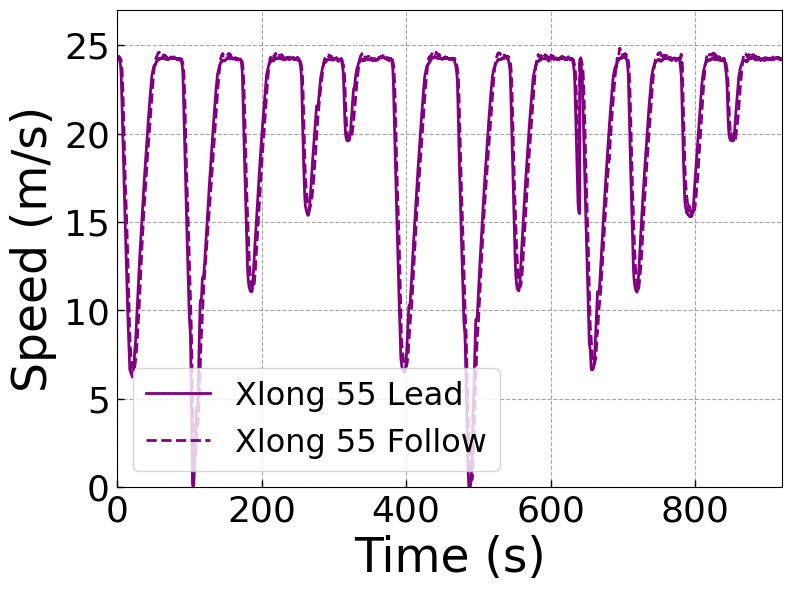}
        \caption{Speed (Xlong)}
        \label{fig:speed_time_xlong_55}
    \end{subfigure}
    
    \vspace{1em}
    \begin{subfigure}{0.24\textwidth}
        \centering
        \includegraphics[width=\textwidth]{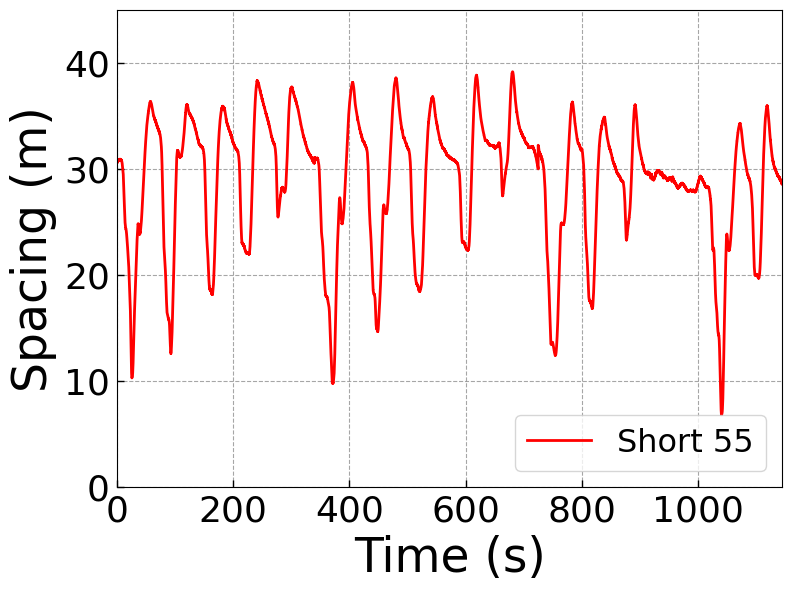}
        \caption{Spacing (Short)}
        \label{fig:spacing_time_short_55}
    \end{subfigure}
    \hfill
    \begin{subfigure}{0.24\textwidth}
        \centering
        \includegraphics[width=\textwidth]{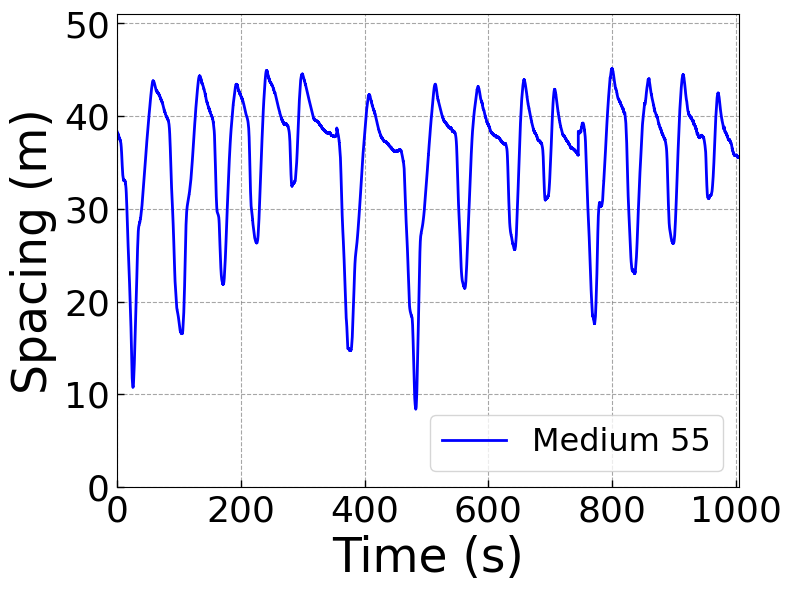}
        \caption{Spacing (Medium)}
        \label{fig:spacing_time_medium_55}
    \end{subfigure}
    \hfill
    \begin{subfigure}{0.24\textwidth}
        \centering
        \includegraphics[width=\textwidth]{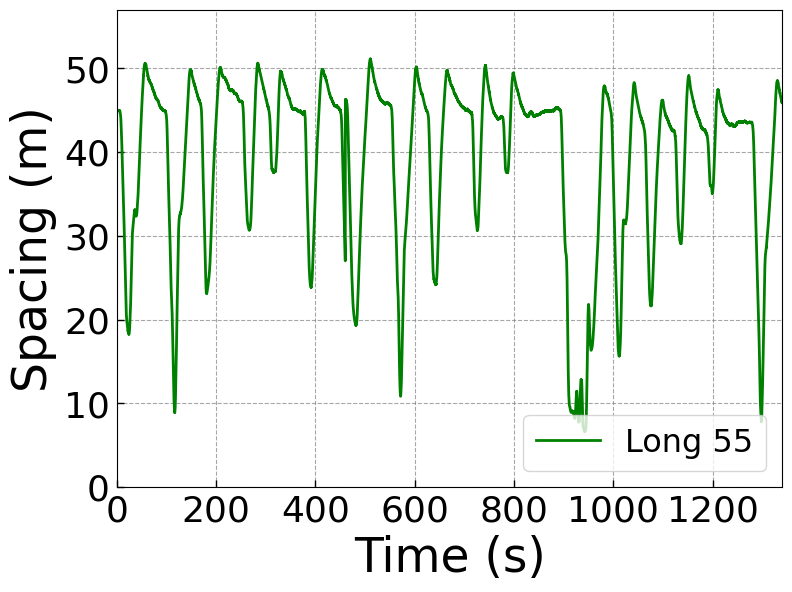}
        \caption{Spacing (Long)}
        \label{fig:spacing_time_long_55}
    \end{subfigure}
    \hfill
    \begin{subfigure}{0.24\textwidth}
        \centering
        \includegraphics[width=\textwidth]{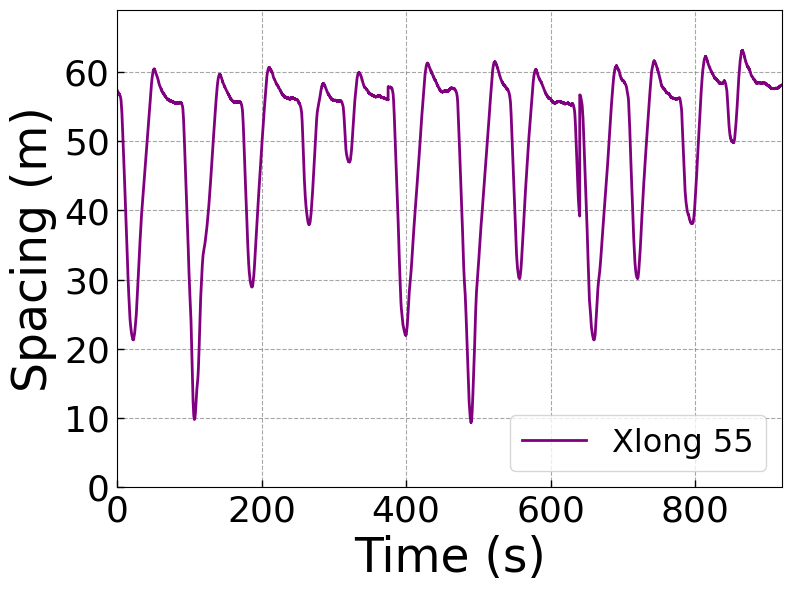}
        \caption{Spacing (Xlong)}
        \label{fig:spacing_time_xlong_55}
    \end{subfigure}
    \caption{Speed and spacing profiles from CF experiments for short (approximately 25~m), medium (approximately 35~m), long (approximately 45~m), and extra-long (approximately 55~m) space-gap settings at 55~mph, representing typical highway free-flow conditions, as defined by initial headway distances.}
    \label{fig:speed_spacing}
\end{figure}

Trajectory data were collected at 25~Hz using a Racebox GPS logger, with spacing computed via the Haversine distance formula. Speed and spacing profiles for the settings are shown in Figure~\ref{fig:speed_spacing}. The input data for the AI-based CF model includes variables such as current spacing \(s\), follower speed \(v\), and speed difference (\(\Delta v\)). Given the limited availability of EV car-following trajectory data, in the future we plan to conduct experiments and share datasets to support EV model research. We intend to make our EV trajectory dataset publicly available, fostering a platform for researchers to enhance data diversity and optimize resource utilization in future experiments.

\subsection{Comparison of ICE and EV Data}

To examine the differences in CF behavior between ICE vehicles and EVs, we analyze key performance metrics using experimental data. Specifically, we investigate acceleration patterns, speed profiles, inter-vehicle spacing, relative speed, and acceleration-deceleration smoothness. To distinguish these differences, we employ comprehensive visualizations, including box plots, probability density functions, Kolmogorov-Smirnov (KS) cumulative distribution functions (CDFs), trajectory plots, and jerk profiles. These metrics highlight the asymmetric CF characteristics of EVs, which are subsequently incorporated into an AI-enabled model to capture their non-symmetric CF behavior. This approach underscores the limitations of conventional ICE-based CF models for EV applications and supports the subsequent comparative analysis of EV dynamics.

\subsubsection{Acceleration Analysis}

\begin{figure}[t!]
\centering
\begin{subfigure}{0.32\textwidth}
    \centering
    \includegraphics[width=\textwidth]{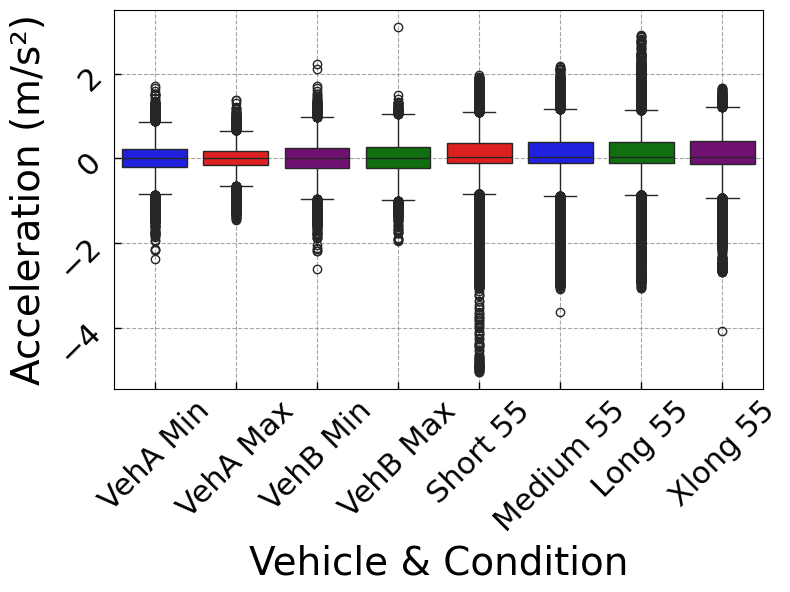}
    \caption{Acceleration box plot.}
    \label{fig:ice_ev_acceleration_boxplot_filtered}
\end{subfigure}
\hfill
\begin{subfigure}{0.32\textwidth}
    \centering
    \includegraphics[width=\textwidth]{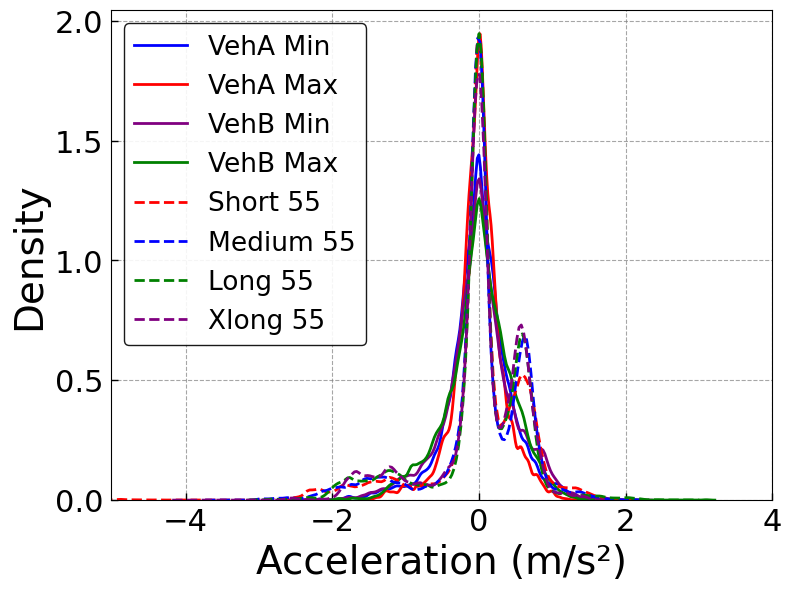}
    \caption{Acceleration density.}
    \label{fig_ice_ev_acc_filtered}
\end{subfigure}
\hfill
\begin{subfigure}{0.32\textwidth}
    \centering
    \includegraphics[width=\textwidth]{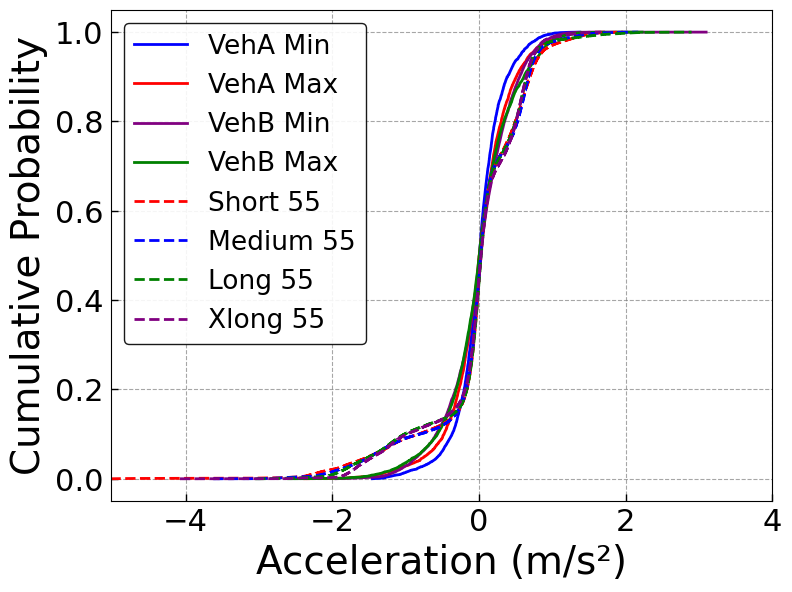}
    \caption{KS-CDF of acceleration.}
    \label{fig:ks_cdf_acceleration_filtered}
\end{subfigure}
\caption{Acceleration comparison between ICE vehicles and EVs.}
\label{fig:acceleration_comparison}
\end{figure}

Figure~\ref{fig:acceleration_comparison} presents an analysis of acceleration distributions for ICE vehicles and EVs through three visualizations: (a) box plots (Figure~\ref{fig:ice_ev_acceleration_boxplot_filtered}), (b) probability density distributions (Figure~\ref{fig_ice_ev_acc_filtered}), and (c) KS-CDFs (Figure~\ref{fig:ks_cdf_acceleration_filtered}). Figure~\ref{fig:ice_ev_acceleration_boxplot_filtered} compares acceleration across configurations. ICE vehicles are shown under minimum and maximum ACC settings (VehA Min, VehA Max, VehB Min, VehB Max), while EV data covers four 55~mph space-gap settings (Short 55, Medium 55, Long 55, Xlong 55). Both types have median accelerations near 0~m/s\(^2\), reflecting steady speeds. The Interquartile Range (IQR) is defined as:
\begin{equation}
\text{IQR} = Q_3 - Q_1
\end{equation}
where \( Q_1 \) and \( Q_3 \) are the first and third quartiles, highlighting differences. EV IQRs narrow at Short 55 and Medium 55, while widening at Long 55 and Xlong 55 due to spacing variability. ICE minimum settings (VehA Min, VehB Min) show wider IQRs than maximum settings, indicating higher variability at smaller gaps.

Whisker length reveals EVs, especially Xlong 55, have a broader acceleration-deceleration range, likely from larger spacing flexibility. ICE medians are centered, suggesting symmetric acceleration-deceleration relationship, while EV Long 55 and Xlong 55 show right-skewed medians, indicating more positive acceleration. Outliers are defined as values outside:
\begin{equation}
[Q_1 - 1.5 \times \text{IQR}, \, Q_3 + 1.5 \times \text{IQR}]
\end{equation}
which appear in EV Xlong 55, suggesting extreme responses.

Table~\ref{tab:acceleration_statistics} quantifies these differences. The median (\( Q_2 \)) is:
\begin{equation}
Q_2 = \text{median}(x_1, x_2, \ldots, x_n)
\end{equation}
with the mean (\(\bar{x}\)) and standard deviation (\(\sigma\)) given by:
\begin{equation}
\bar{x} = \frac{1}{n} \sum_{i=1}^n x_i
\end{equation}
and
\begin{equation}
\sigma = \sqrt{\frac{1}{n} \sum_{i=1}^n (x_i - \bar{x})^2}
\end{equation}
where \( n \) is the sample size and \( x_i \) are acceleration values. EVs show higher variability (e.g., \(\sigma = 0.675\)~m/s\textsuperscript{2} for Short 55) than ICE vehicles (e.g., \(\sigma = 0.337\)~m/s\textsuperscript{2} for VehA Max), supporting vehicle-specific modeling.

\begin{table}[t!]
\captionsetup{justification=centering}
\vspace{1em}
\caption{Acceleration statistics for ICE vehicles and EVs (in \si{m/s^2}).}
\begin{center}
\setlength\tabcolsep{3.5pt}
\begin{tabular}{c c *{5}{S[table-format=3.3, round-mode=places, round-precision=3]}}
\toprule
\textbf{Vehicle Type} & \textbf{Vehicle Info} & \textbf{Max} & \textbf{Min} & \textbf{Median} & \textbf{Mean} & \textbf{Std} \\
\midrule
\multirow{4}{*}{ICE}
& VehA Min & 1.699 & -2.372 & 0.003 & 0.000 & 0.438 \\
& VehA Max & 1.375 & -1.453 & 0.005 & -0.000 & 0.337 \\
& VehB Min & 2.224 & -2.595 & 0.004 & -0.000 & 0.484 \\
& VehB Max & 3.095 & -1.951 & 0.012 & -0.001 & 0.454 \\
\midrule
\multirow{4}{*}{EV}
& Short 55 & 1.964 & -5.022 & 0.028 & -0.002 & 0.675 \\
& Medium 55 & 2.169 & -3.622 & 0.025 & -0.000 & 0.644 \\
& Long 55 & 2.897 & -3.050 & 0.031 & -0.007 & 0.649 \\
& Xlong 55 & 1.667 & -4.075 & 0.028 & -0.009 & 0.609 \\
\bottomrule
\end{tabular}
\end{center}
\vspace{-20pt}
\label{tab:acceleration_statistics}
\end{table}

The probability density distribution (Figure~\ref{fig_ice_ev_acc_filtered}), computed via kernel density estimation (KDE), is given by:
\begin{equation}
\hat{f}_h(x) = \frac{1}{nh} \sum_{i=1}^n K\left(\frac{x - x_i}{h}\right)
\end{equation}
where \(\hat{f}_h(x)\) is the density, \( n \) is the number of data points, \( h \) is bandwidth, and \( K \) is the Gaussian kernel, revealing EV multimodal acceleration patterns. The EV distributions exhibit a primary peak near 0~m/s\(^2\), with secondary peaks at 0.5 to 1.5~m/s\(^2\) and -2.5 to -1.5~m/s\(^2\) in acceleration and deceleration phases, as shown in Figure~\ref{fig_ice_ev_acc_filtered}. These peaks reflect EV-ICE differences, with faster EV motor responses driving acceleration at 0.5 to 1.5~m/s\(^2\) and regenerative braking introducing additional deceleration at -2.5 to -1.5~m/s\(^2\).

Figure~\ref{fig:ks_cdf_acceleration_filtered} compares CDFs, showing S-shaped curves with a steep gradient near 0~m/s\(^2\). EV Xlong 55 rises above ICE in the 0.5 to 1.5~m/s\(^2\) range, and Long 55 and Xlong 55 steepen at -2.5 to -1.5~m/s\(^2\), confirming enhanced EV dynamics. This asymmetry challenges traditional CF models, driven by powertrain differences rather than experimental variations.

\subsubsection{Relative Speed Analysis}

\begin{figure}[t!]
    \centering
    \begin{subfigure}{0.32\textwidth}
        \centering
        \includegraphics[width=\textwidth]{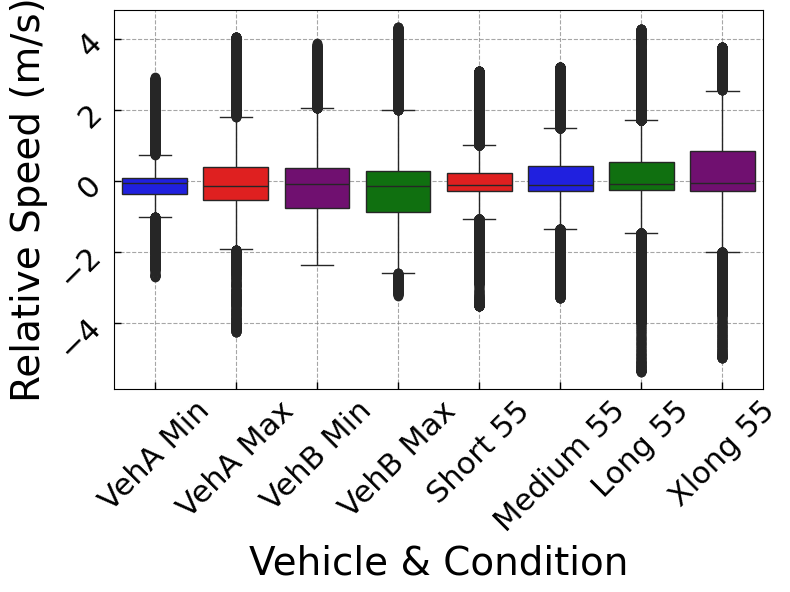}
        \caption{Relative speed box plot.}
        \label{fig:ice_ev_rel_speed_boxplot_filtered}
    \end{subfigure}
    \hfill
    \begin{subfigure}{0.32\textwidth}
        \centering
        \includegraphics[width=\textwidth]{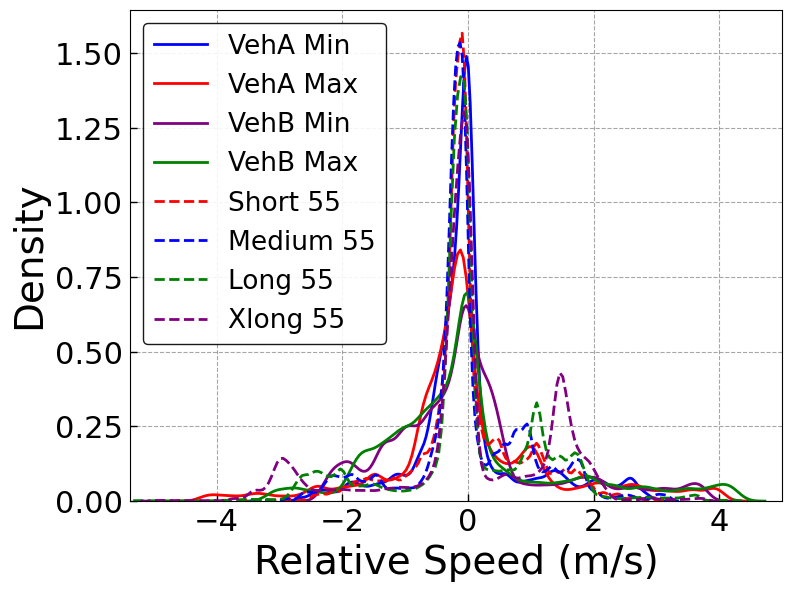}
        \caption{Relative speed density.}
        \label{fig:ice_ev_rel_speed_density_filtered}
    \end{subfigure}
    \hfill
    \begin{subfigure}{0.32\textwidth}
        \centering
        \includegraphics[width=\textwidth]{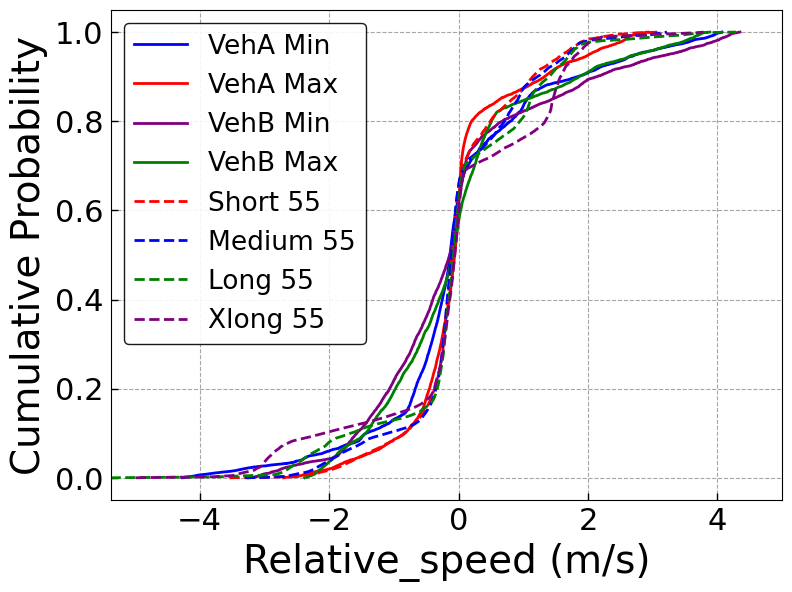}
        \caption{KS-CDF of relative speed.}
        \label{fig:ks_cdf_rel_speed_filtered}
    \end{subfigure}
    \caption{Relative speed comparison for ICE vehicles and EVs: (a) box plot revealing gap-dependent variability, (b) probability density distributions showing asymmetric EV patterns at larger gaps, and (c) KS-CDF plot confirming systematic behavioral differences between vehicle types.}
    \label{fig:rel_speed_comparison}
\end{figure}

Figure~\ref{fig:rel_speed_comparison} presents relative speed characteristics for ICE vehicles and EVs through three visualizations: (a) box plots (Figure~\ref{fig:ice_ev_rel_speed_boxplot_filtered}), (b) probability density distributions (Figure~\ref{fig:ice_ev_rel_speed_density_filtered}), and (c) KS-CDFs (Figure~\ref{fig:ks_cdf_rel_speed_filtered}). Figure~\ref{fig:ice_ev_rel_speed_boxplot_filtered} highlights speed-matching differences. Both ICE vehicles and EVs show median relative speeds near 0~m/s, indicating alignment with the lead vehicle. IQRs reveal distinct patterns. Specifically, ICE IQRs are consistent, with minimum settings (VehA Min, VehB Min) being slightly wider than maximum settings (VehA Max, VehB Max), suggesting variability at smaller gaps. EV IQRs increase from Short 55 to Xlong 55, reflecting flexibility at larger spacings. EV Long 55 and Xlong 55 show right-skewed medians, indicating a tendency for higher speeds.

Table~\ref{tab:relative_speed_statistics} quantifies these findings. EV relative speeds at larger gaps (e.g., \(\sigma = 1.299\)~m/s for Xlong 55) show higher variability than ICE vehicles (e.g., \(\sigma = 0.902\)~m/s for VehA Min), reflecting dynamic adjustments by electric powertrains. The probability density distribution (Figure~\ref{fig:ice_ev_rel_speed_density_filtered}) shows ICE vehicles with symmetric peaks at 0~m/s, while EV distributions become asymmetric at Long 55 and Xlong 55, with peaks at +0.3 to +0.7~m/s and -2.5 to -1.5~m/s, due to rapid acceleration and regenerative braking.

The KS-CDF analysis presented in Figure~\ref{fig:ks_cdf_rel_speed_filtered} confirms these patterns. All vehicles show S-shaped CDFs with steep gradients near 0~m/s. ICE CDFs are symmetric, while EV Long 55 and Xlong 55 diverge, with steeper ascents at -1.5 to 0~m/s and gradual rises at 0 to 0.7~m/s, highlighting dynamic EV speed profiles. This asymmetry challenges conventional CF models and necessitates tailored approaches to characterize the car-following dynamics of EVs.

\begin{table}[t!]
\captionsetup{justification=centering}
\vspace{1em}
\caption{Relative speed statistics for ICE vehicles and EVs (in \si{m/s}).}
\begin{center}
\setlength\tabcolsep{3.5pt}
\begin{tabular}{c c *{5}{S[table-format=3.3, round-mode=places, round-precision=3]}}
\toprule
\textbf{Vehicle Type} & \textbf{Vehicle Info} & \textbf{Max} & \textbf{Min} & \textbf{Median} & \textbf{Mean} & \textbf{Std} \\
\midrule
\multirow{4}{*}{ICE}
& VehA Min & 2.926 & -2.701 & -0.059 & 0.001 & 0.902 \\
& VehA Max & 4.071 & -4.265 & -0.133 & 0.001 & 1.350 \\
& VehB Min & 3.883 & -2.378 & -0.093 & -0.011 & 1.253 \\
& VehB Max & 4.347 & -3.242 & -0.137 & -0.016 & 1.448 \\
\midrule
\multirow{4}{*}{EV}
& Short 55 & 3.110 & -3.539 & -0.100 & -0.003 & 0.828 \\
& Medium 55 & 3.219 & -3.299 & -0.109 & -0.001 & 0.932 \\
& Long 55 & 4.289 & -5.377 & -0.079 & -0.010 & 1.126 \\
& Xlong 55 & 3.784 & -4.985 & -0.058 & -0.018 & 1.299 \\
\bottomrule
\end{tabular}
\end{center}
\label{tab:relative_speed_statistics}
\end{table}

\subsubsection{Speed Analysis}

\begin{figure}[t!]
    \centering
    \begin{subfigure}{0.32\textwidth}
        \centering
        \includegraphics[width=\textwidth]{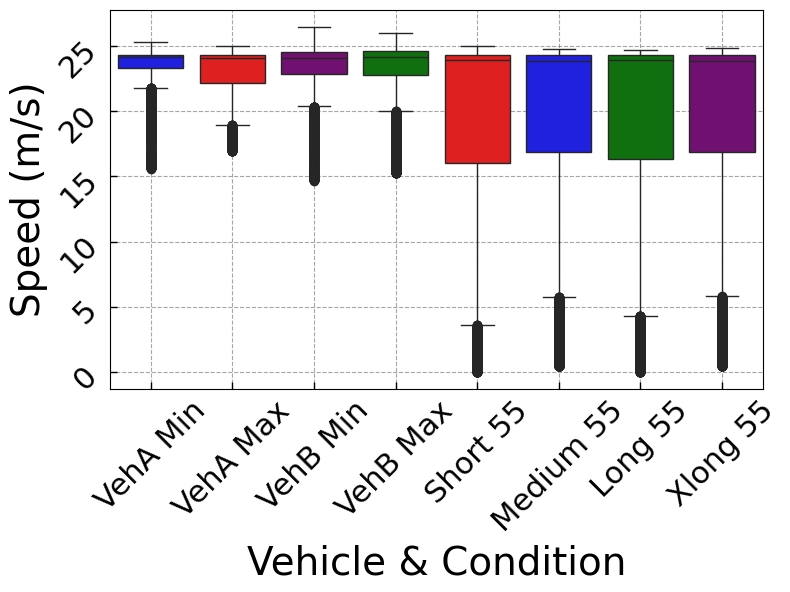}
        \caption{Speed box plot.}
        \label{fig:ice_ev_speed_boxplot}
    \end{subfigure}
    \hfill
    \begin{subfigure}{0.32\textwidth}
        \centering
        \includegraphics[width=\textwidth]{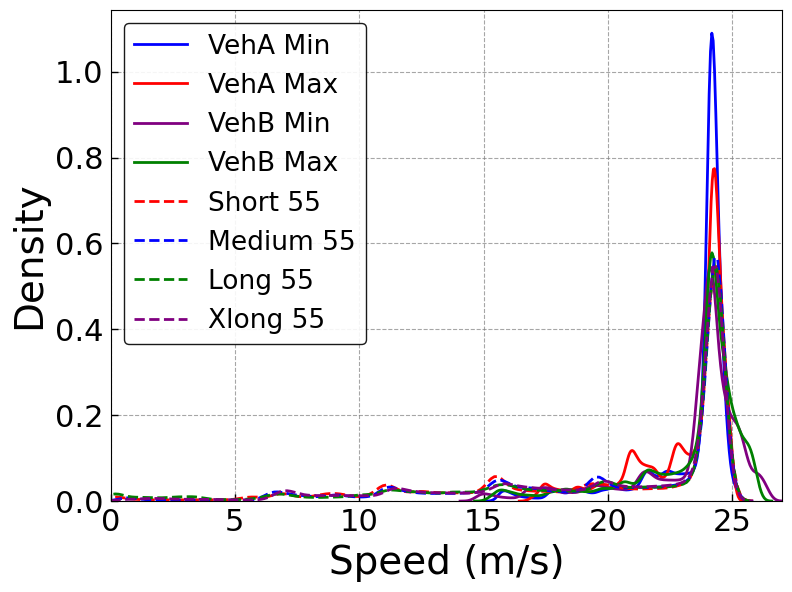}
        \caption{Speed density.}
        \label{fig:ice_ev_speed_density}
    \end{subfigure}
    \hfill
    \begin{subfigure}{0.32\textwidth}
        \centering
        \includegraphics[width=\textwidth]{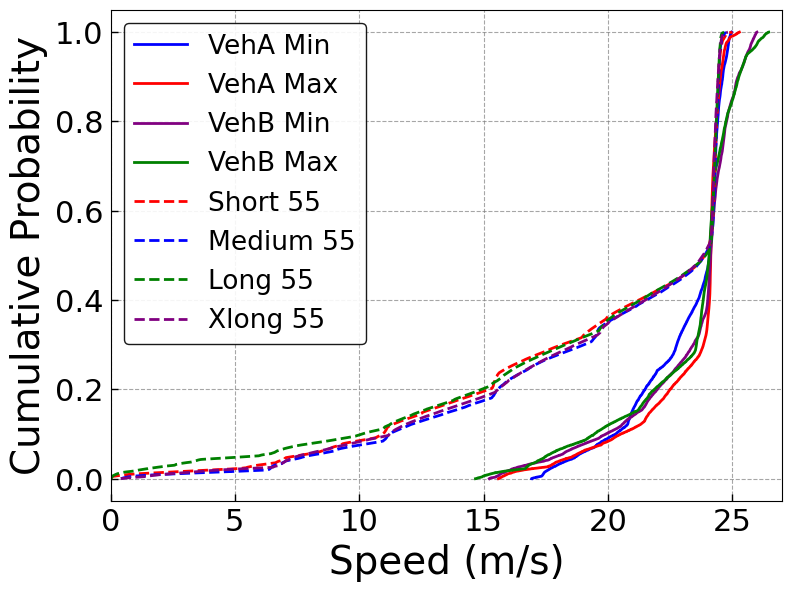}
        \caption{KS-CDF of speed.}
        \label{fig:ks_cdf_speed}
    \end{subfigure}
    \caption{Speed comparison for ICE vehicles and EVs: (a) box plot showing conservative ICE behavior versus precise EV speed maintenance, (b) probability density distributions revealing concentrated EV peaks at target speed, and (c) KS-CDF plot quantifying the distinct speed-maintenance strategies.}
    \label{fig:speed_comparison}
\end{figure}

Figure~\ref{fig:speed_comparison} examines speed maintenance for ICE vehicles and EVs through three visualizations: (a) box plots (Figure~\ref{fig:ice_ev_speed_boxplot}), (b) probability density distributions (Figure~\ref{fig:ice_ev_speed_density}), and (c) KS-CDFs (Figure~\ref{fig:ks_cdf_speed}). Figure~\ref{fig:ice_ev_speed_boxplot} highlights vehicle speed strategies. ICE vehicles show median speeds of 23.5–24~m/s with narrow IQRs, indicating conservative behavior below the 25–26~m/s target. Minimum ACC settings (VehA Min, VehB Min) have wider IQRs than maximum settings, suggesting variability at smaller gaps. EVs cluster a median speed of 24.6~m/s (55~mph), with Short 55 and Medium 55 exhibiting narrow IQRs, while Long 55 and Xlong 55 show slightly wider IQRs, reflecting precise control with minor spacing variations.

Table~\ref{tab:speed_statistics} quantifies this contrast. ICE vehicles have low variability (e.g., \(\sigma = 1.886\)~m/s for VehA Max), while EVs show higher variability (e.g., \(\sigma = 5.972\)~m/s for Short 55), due to transient maneuvers. The probability density distribution (Figure~\ref{fig:ice_ev_speed_density}) shows ICE vehicle curves with broad peaks at 23.5–24~m/s, confirming conservatism. EV curves peak sharply at 24.6~m/s, with Long 55 and Xlong 55 broadening, reflecting aggressive control via torque response.

The KS-CDF analysis presented in Figure~\ref{fig:ks_cdf_speed} confirms differences in the CDFs between ICE vehicles and EVs. The CDFs of ICE vehicles plateau around 23.5–24~m/s, indicating consistent speed behavior. In contrast, EV CDFs rise steeply at 24.6~m/s, with Long 55 and Xlong 55 showing less steep increases, reflecting broader speed distributions. This level of precision challenges traditional CF models and underscores the need for EV-specific modeling approaches.

\begin{table}[t!]
\captionsetup{justification=centering}
\vspace{1em}
\caption{Speed statistics for ICE vehicles and EVs (in \si{m/s}).}
\begin{center}
\setlength\tabcolsep{3.5pt}
\begin{tabular}{c c *{5}{S[table-format=3.3, round-mode=places, round-precision=3]}}
\toprule
\textbf{Vehicle Type} & \textbf{Vehicle Info} & \textbf{Max} & \textbf{Min} & \textbf{Median} & \textbf{Mean} & \textbf{Std} \\
\midrule
\multirow{4}{*}{ICE}
& VehA Min & 25.278 & 15.595 & 24.129 & 23.315 & 1.918 \\
& VehA Max & 24.965 & 16.923 & 24.077 & 23.081 & 1.886 \\
& VehB Min & 26.458 & 14.671 & 24.061 & 23.238 & 2.368 \\
& VehB Max & 25.984 & 15.226 & 24.119 & 23.315 & 2.269 \\
\midrule
\multirow{4}{*}{EV}
& Short 55 & 25.004 & 0.023 & 23.888 & 19.998 & 5.972 \\
& Medium 55 & 24.755 & 0.433 & 23.884 & 20.291 & 5.654 \\
& Long 55 & 24.665 & 0.022 & 23.914 & 19.812 & 6.418 \\
& Xlong 55 & 24.817 & 0.414 & 23.811 & 20.185 & 5.755 \\
\bottomrule
\end{tabular}
\end{center}
\label{tab:speed_statistics}
\end{table}

\subsubsection{Spacing Analysis}

\begin{figure}[t!]
    \centering
    \begin{subfigure}{0.32\textwidth}
        \centering
        \includegraphics[width=\textwidth]{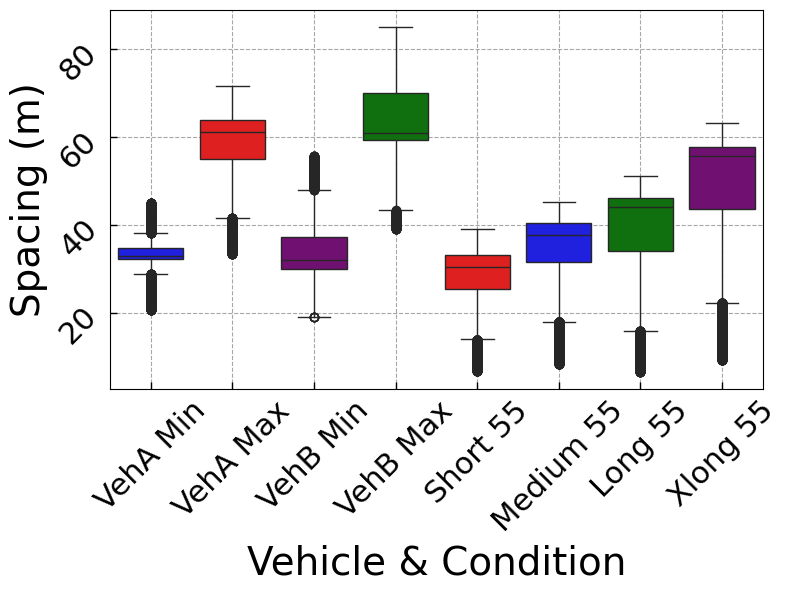}
        \caption{Spacing box plot.}
        \label{fig:ice_ev_spacing_boxplot}
    \end{subfigure}
    \hfill
    \begin{subfigure}{0.32\textwidth}
        \centering
        \includegraphics[width=\textwidth]{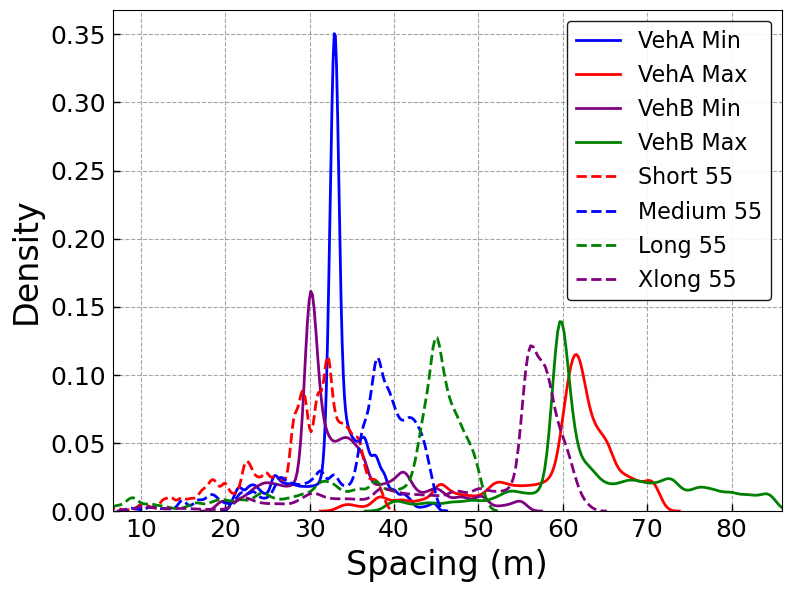}
        \caption{Spacing density.}
        \label{fig:ice_ev_spacing_density}
    \end{subfigure}
    \hfill
    \begin{subfigure}{0.32\textwidth}
        \centering
        \includegraphics[width=\textwidth]{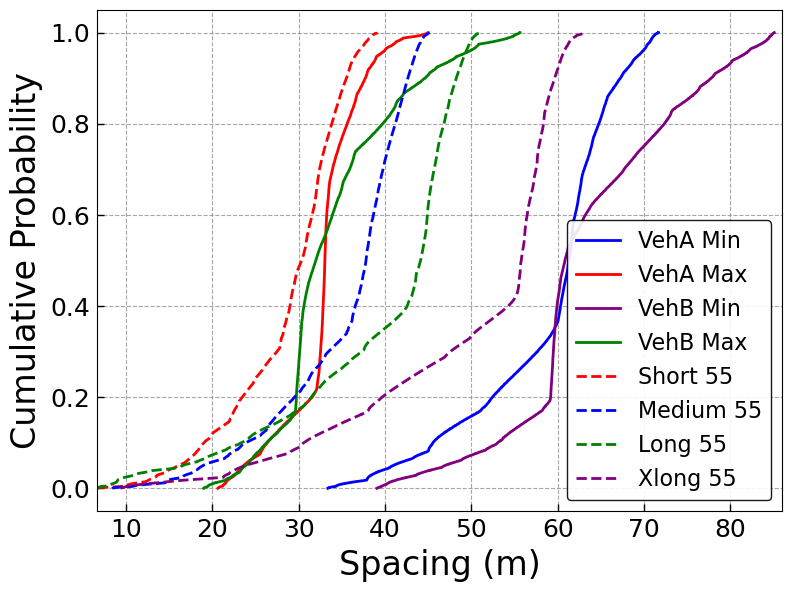}
        \caption{KS-CDF for spacing.}
        \label{fig:ks_cdf_spacing}
    \end{subfigure}
    \caption{Spacing comparison for ICE vehicles and EVs: (a) box plot revealing systematic gap-dependent EV spacing, (b) probability density distributions showing distinct peaks for each EV gap setting, and (c) KS-CDF plot confirming fundamentally different spacing strategies between vehicle types.}
    \label{fig:spacing_comparison}
\end{figure}

Figure~\ref{fig:spacing_comparison} investigates gap-maintenance characteristics for ICE vehicles and EVs through three visualizations: (a) box plots (Figure~\ref{fig:ice_ev_spacing_boxplot}), (b) probability density distributions (Figure~\ref{fig:ice_ev_spacing_density}), and (c) KS-CDFs (Figure~\ref{fig:ks_cdf_spacing}). Figure~\ref{fig:ice_ev_spacing_boxplot} highlights spacing strategies. ICE vehicles maintain fixed medians of 45–50 m (VehA Min, VehB Min) and 60–65~m (VehA Max, VehB Max), with moderate IQRs reflecting mechanical constraints. EVs show gap-dependent medians of 25~m (Short 55), 35~m (Medium 55), 45~m (Long 55), and 55~m (Xlong 55), with IQRs widening at larger gaps, indicating dynamic adjustments via rapid response capabilities.

Table~\ref{tab:spacing_statistics} quantifies these differences. Specifically, ICE vehicle variability is consistent (e.g., \(\sigma = 4.193\) m for VehA Min), while EV standard deviations increase with gap size (e.g., \(\sigma = 11.824\) m for Xlong 55), reflecting dynamic spacing. The probability density distribution (Figure~\ref{fig:ice_ev_spacing_density}) shows ICE vehicle curves as normal with narrow spreads at ACC settings. However, EV curves peak at 25~m, 35~m, 45~m, and 55~m, broadening at larger gaps, leveraging torque and braking for flexibility.

The KS-CDF analysis shown in Figure~\ref{fig:ks_cdf_spacing} further validates these patterns. The CDFs of ICE vehicles plateau at ACC settings, indicating consistent spacing behavior. In contrast, EV CDFs exhibit step-like increases at nominal spacings, with Xlong 55 peaking at 55~m, suggesting deliberate and precise gap management. These dynamic spacing patterns among EVs challenge traditional CF models and highlight the need for new paradigms to capture enhanced maneuvering behavior.

\begin{table}[t!]
\captionsetup{justification=centering}
\vspace{1em}
\caption{Spacing statistics for ICE vehicles and EVs (in \si{m}).}
\begin{center}
\setlength\tabcolsep{3.5pt}
\begin{tabular}{c c *{5}{S[table-format=3.3, round-mode=places, round-precision=3]}}
\toprule
\textbf{Vehicle Type} & \textbf{Vehicle Info} & \textbf{Max} & \textbf{Min} & \textbf{Median} & \textbf{Mean} & \textbf{Std} \\
\midrule
\multirow{4}{*}{ICE}
& VehA Min & 45.023 & 20.667 & 33.052 & 32.855 & 4.193 \\
& VehA Max & 71.668 & 33.393 & 61.284 & 58.847 & 8.165 \\
& VehB Min & 55.643 & 19.013 & 32.002 & 33.991 & 7.089 \\
& VehB Max & 85.078 & 39.068 & 60.844 & 63.589 & 9.540 \\
\midrule
\multirow{4}{*}{EV}
& Short 55 & 39.157 & 6.817 & 30.368 & 28.798 & 6.249 \\
& Medium 55 & 45.139 & 8.418 & 37.812 & 35.349 & 7.373 \\
& Long 55 & 51.140 & 6.629 & 44.135 & 39.374 & 10.294 \\
& Xlong 55 & 63.137 & 9.332 & 55.784 & 49.936 & 11.824 \\
\bottomrule
\end{tabular}
\end{center}
\label{tab:spacing_statistics}
\end{table}

\subsubsection{Analysis of Acceleration and Deceleration Smoothness}

\begin{figure}[t!]
    \centering
    \begin{subfigure}{0.24\textwidth}
        \centering
        \includegraphics[width=\textwidth]{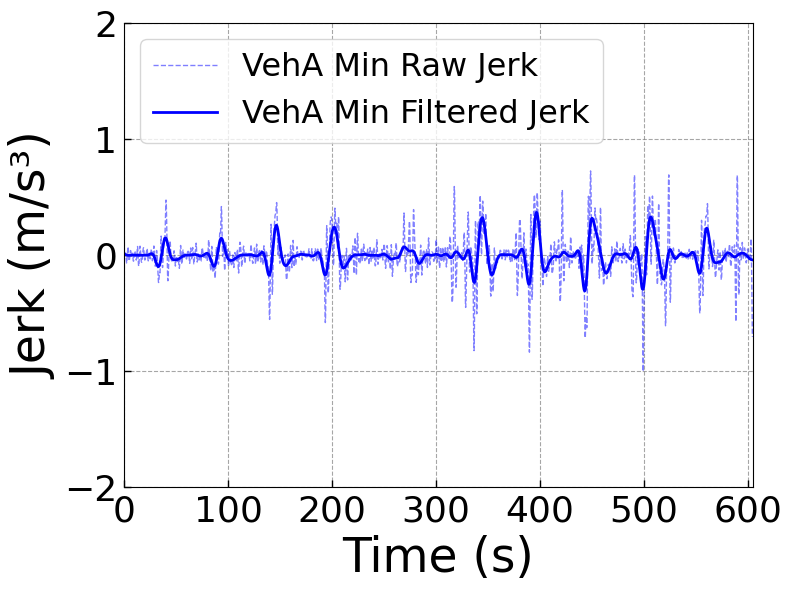}
        \caption{ICE VehA Min jerk.}
        \label{fig:ice_VehA Min_jerk}
    \end{subfigure}
    \hfill
    \begin{subfigure}{0.24\textwidth}
        \centering
        \includegraphics[width=\textwidth]{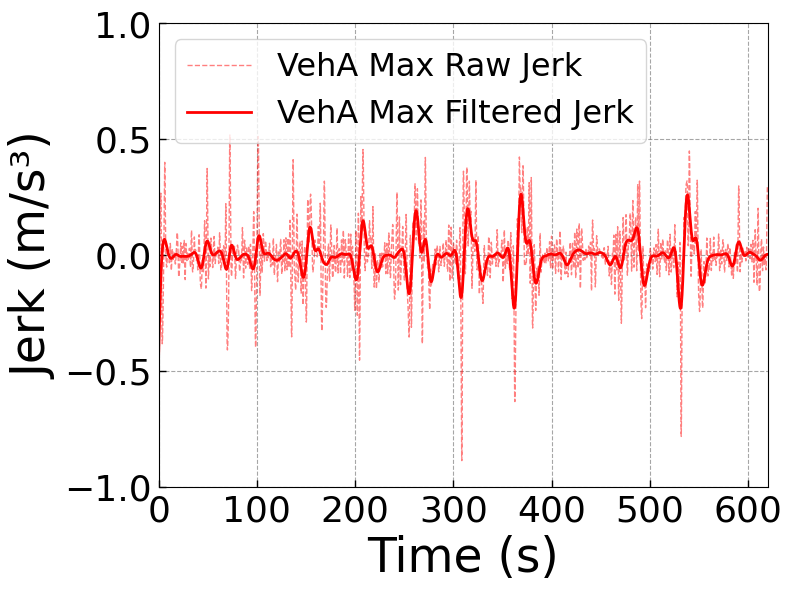}
        \caption{ICE VehA Max jerk.}
        \label{fig:ice_vehA_max_jerk}
    \end{subfigure}
    \hfill
    \begin{subfigure}{0.24\textwidth}
        \centering
        \includegraphics[width=\textwidth]{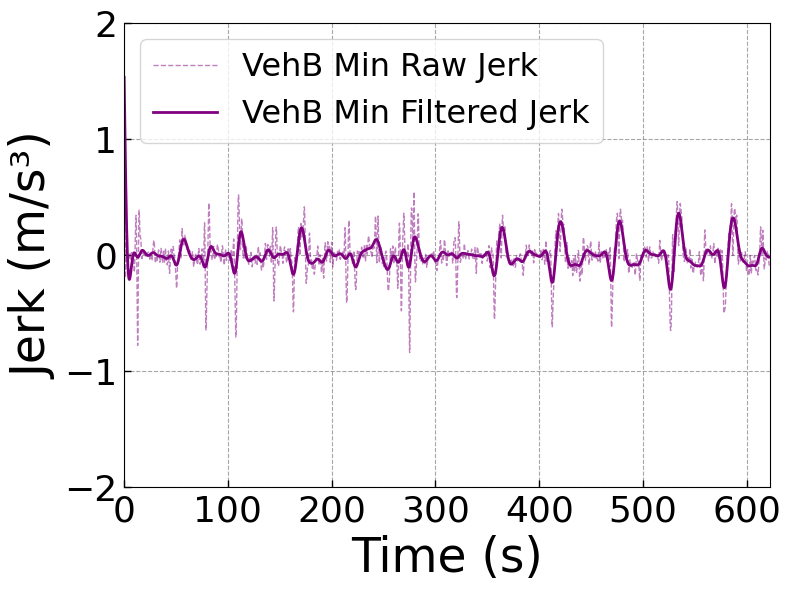}
        \caption{ICE VehB Min jerk.}
        \label{fig:ice_vehB_min_jerk}
    \end{subfigure}
    \hfill
    \begin{subfigure}{0.24\textwidth}
        \centering
        \includegraphics[width=\textwidth]{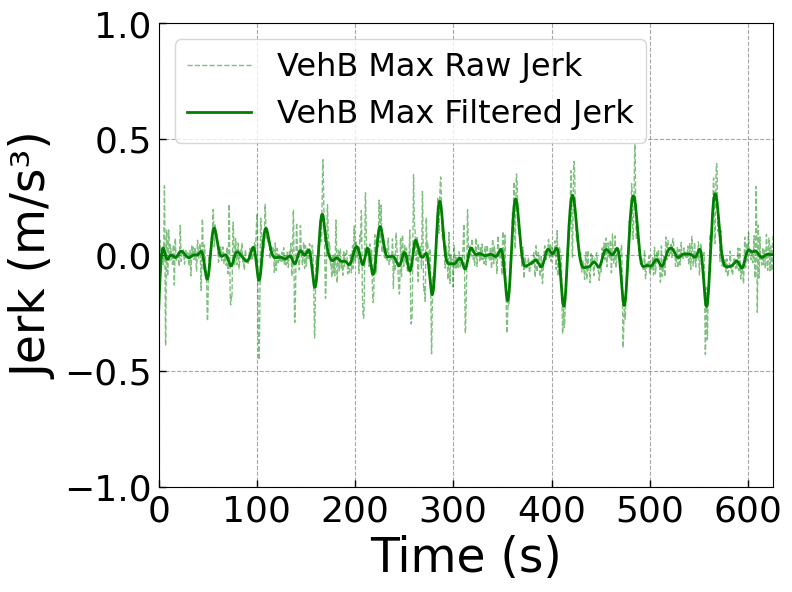}
        \caption{ICE VehB Max jerk.}
        \label{fig:ice_vehB_max_jerk}
    \end{subfigure}
    
    \vspace{1em}
    \begin{subfigure}{0.24\textwidth}
        \centering
        \includegraphics[width=\textwidth]{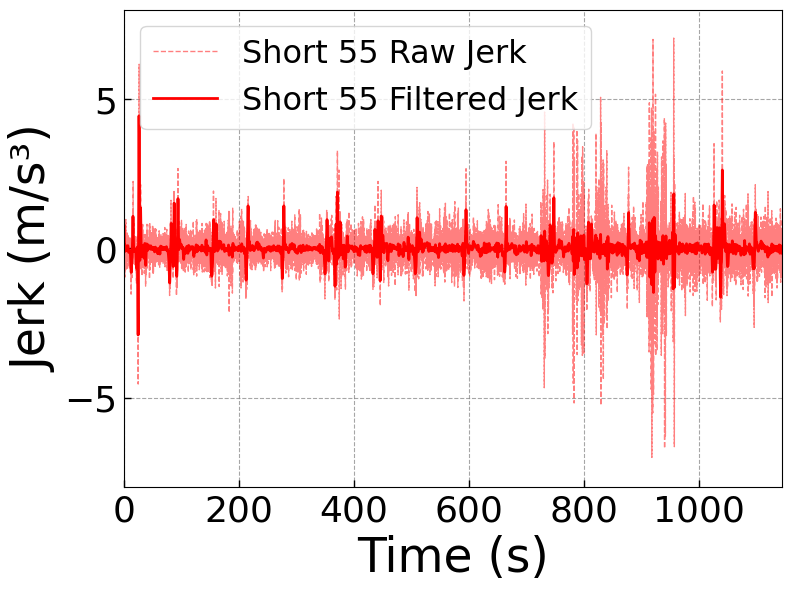}
        \caption{EV Short 55 jerk.}
        \label{fig:ev_short_55_jerk}
    \end{subfigure}
    \hfill
    \begin{subfigure}{0.24\textwidth}
        \centering
        \includegraphics[width=\textwidth]{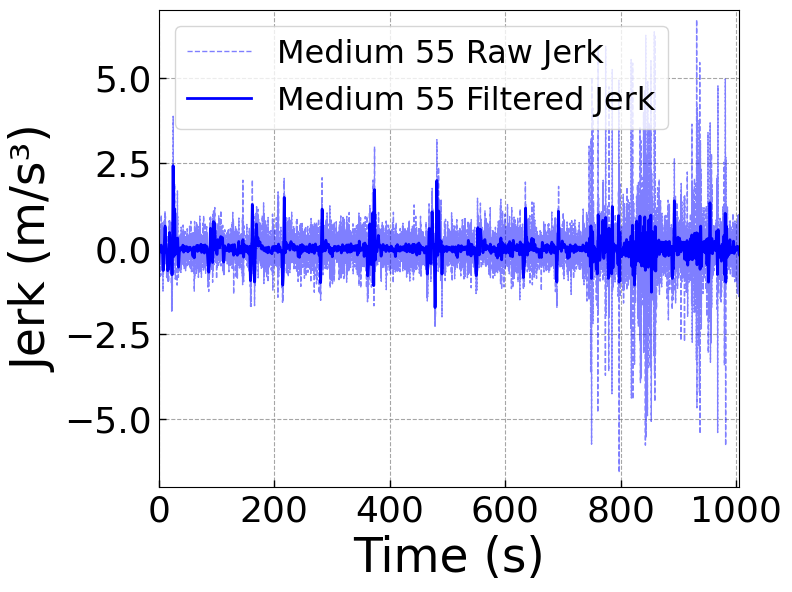}
        \caption{EV Medium 55 jerk.}
        \label{fig:ev_medium_55_jerk}
    \end{subfigure}
    \hfill
    \begin{subfigure}{0.24\textwidth}
        \centering
        \includegraphics[width=\textwidth]{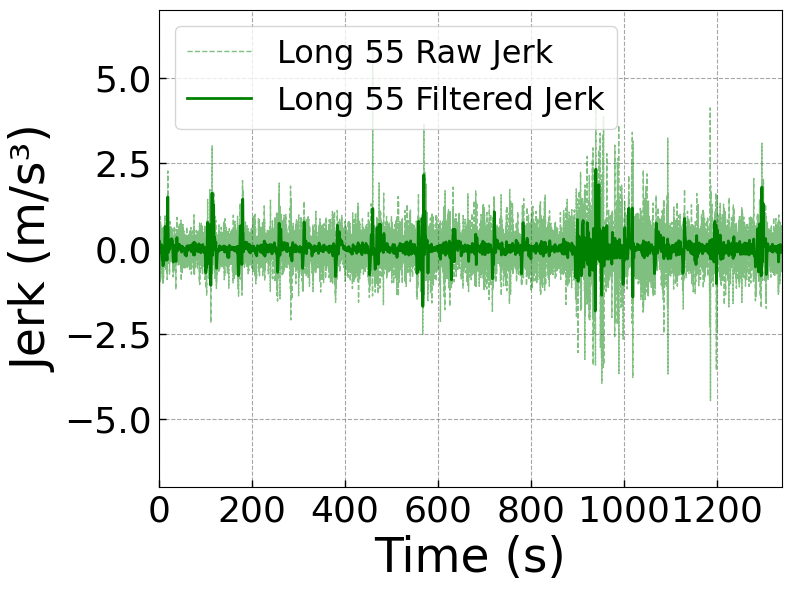}
        \caption{EV Long 55 jerk.}
        \label{fig:ev_long_55_jerk}
    \end{subfigure}
    \hfill
    \begin{subfigure}{0.24\textwidth}
        \centering
        \includegraphics[width=\textwidth]{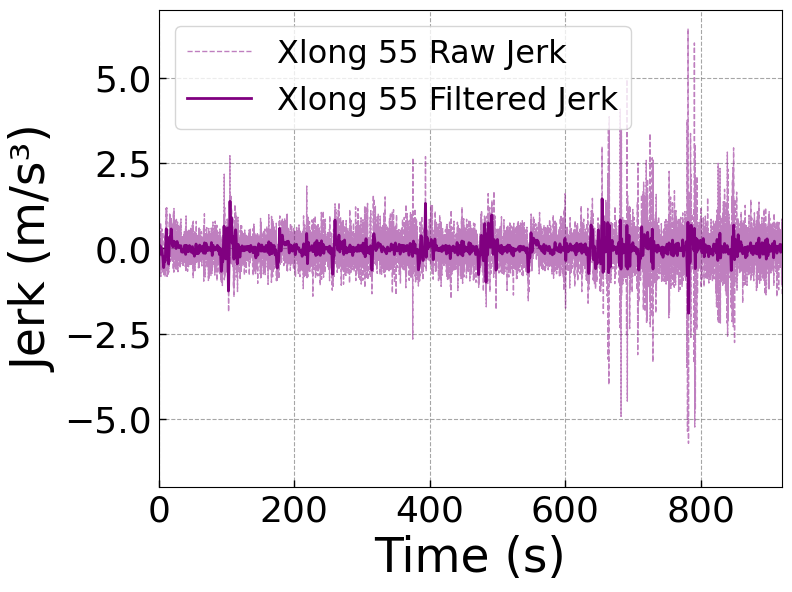}
        \caption{EV Xlong 55 jerk.}
        \label{fig:ev_xlong_55_jerk}
    \end{subfigure}
    \caption{Jerk profiles for ICE vehicles and EVs, with the first row showing ICE vehicles (VehA Min, VehA Max, VehB Min, VehB Max) and the second row showing EV conditions (Short 55, Medium 55, Long 55, Xlong 55), highlighting the smoother acceleration and deceleration behavior of EVs compared to ICE vehicles.}
    \label{fig:jerk_comparison}
\end{figure}

The smoothness of acceleration and deceleration for ICE vehicles and EVs is analyzed using jerk profiles, as illustrated in Figure~\ref{fig:jerk_comparison}. Jerk, defined as the rate of change of acceleration, is calculated as follows:
\begin{equation}
j(t) = \frac{\Delta a(t)}{\Delta t} = \frac{a(t) - a(t-1)}{t - (t-1)}
\end{equation}
where \( j(t) \) is jerk at time \( t \), \( a(t) \) is acceleration, and \( \Delta t \) is the time interval. Figure~\ref{fig:jerk_comparison} displays raw (dashed) and filtered (solid) jerk for ICE (VehA Min, VehA Max, VehB Min, VehB Max) and EV (Short 55, Medium 55, Long 55, Xlong 55) conditions.

\begin{table}[t!]
\captionsetup{justification=centering}
\vspace{1em}
\caption{Jerk statistics for ICE vehicles and EVs (in \si{m/s^3} for Max, Min, Median, Mean, Std; \si{m^2/s^6} for Squared Integral).}
\begin{center}
\setlength\tabcolsep{3.5pt}
\begin{tabular}{c c *{6}{S[table-format=3.3, round-mode=places, round-precision=3]}}
\toprule
\textbf{Vehicle Type} & \textbf{Vehicle Info} & \textbf{Max} & \textbf{Min} & \textbf{Median} & \textbf{Mean} & \textbf{Std} & \textbf{Sq. Integral} \\
\midrule
\multirow{4}{*}{ICE}
& VehA Min & 0.369 & -0.308 & -0.001 & -0.000 & 0.095 & 5.478 \\
& VehA Max & 0.263 & -0.429 & 0.001 & -0.001 & 0.065 & 2.608 \\
& VehB Min & 1.534 & -0.295 & -0.004 & 0.003 & 0.110 & 7.237 \\
& VehB Max & 0.265 & -0.239 & -0.005 & -0.000 & 0.074 & 3.375 \\
\midrule
\multirow{4}{*}{EV}
& Short 55 & 4.447 & -2.901 & -0.003 & 0.000 & 0.332 & 126.228 \\
& Medium 55 & 2.427 & -1.734 & -0.004 & -0.000 & 0.293 & 86.503 \\
& Long 55 & 2.328 & -1.834 & -0.005 & -0.000 & 0.264 & 93.638 \\
& Xlong 55 & 1.458 & -1.904 & -0.004 & 0.000 & 0.219 & 44.059 \\
\bottomrule
\end{tabular}
\end{center}
\label{tab:jerk_statistics}
\end{table}

The jerk profiles of ICE vehicles show modest fluctuations, with raw and filtered values ranging from -4 to 4~m/s\(^3\) and -0.8 to 0.8~m/s\(^3\), respectively. The minimum settings exhibit higher-frequency fluctuations compared to the maximum settings, indicating that larger gaps result in smoother responses. EV jerk profiles exhibit more intense behavior, with raw values from -10 to 10~m/s\(^3\) and filtered values from -2 to 2~m/s\(^3\). The Xlong 55 condition shows the largest fluctuations, reflecting rapid torque and regenerative braking at larger spacings.

Table~\ref{tab:jerk_statistics} presents statistical metrics, including the jerk squared integral (JSI), defined as:
\begin{equation}
\text{JSI} = \int_{t_1}^{t_2} j(t)^2 \, dt
\end{equation}
where \( \text{JSI} \) quantifies the cumulative intensity of jerk fluctuations over the interval \([t_1, t_2]\). Greater JSI values observed in EVs (e.g., 126.228 m\(^2\)/s\(^6\) for Short 55) compared to ICE vehicles (e.g., 5.478 m\(^2\)/s\(^6\) for VehA Min) indicate reduced smoothness, supporting the inclusion of jerk dynamics as distinguishing features in AI-based CF models. These differences underscore the more intense CF behavior of EVs and further challenge the assumptions of traditional CF models.

\subsubsection{Speed-Spacing Trajectory Analysis}

\begin{figure}[t!]
    \centering
    \begin{subfigure}{0.45\textwidth}
        \centering
        \includegraphics[width=\textwidth]{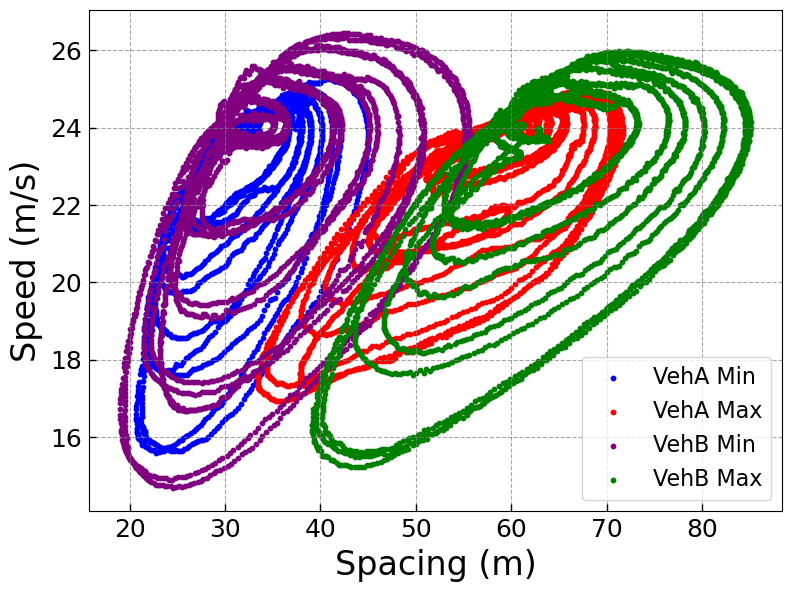}
        \caption{ICE speed-spacing trajectories.}
        \label{fig:ice_speed_spacing}
    \end{subfigure}
    \hfill
    \begin{subfigure}{0.45\textwidth}
        \centering
        \includegraphics[width=\textwidth]{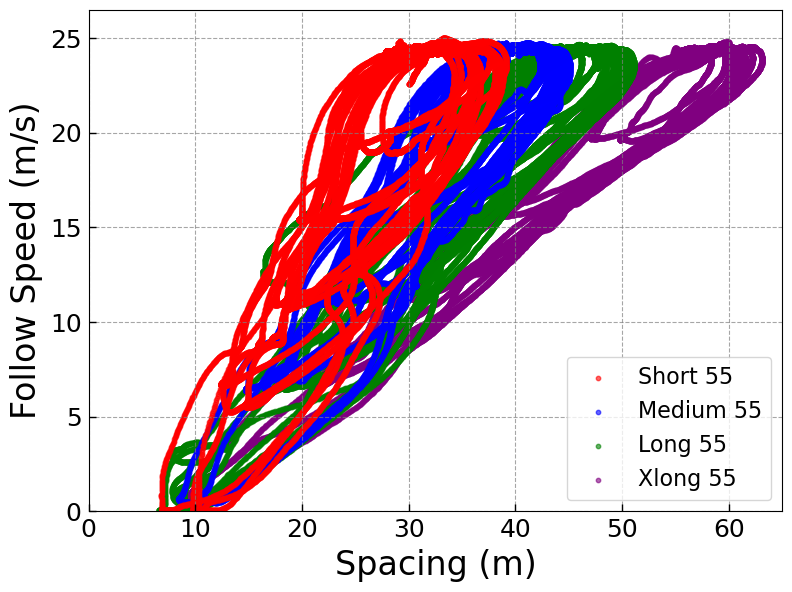}
        \caption{EV speed-spacing trajectories.}
        \label{fig:ev_speed_spacing}
    \end{subfigure}
    \caption{Speed-spacing trajectories for ICE vehicles and EVs, demonstrating the near-linear correlation in EVs versus the annular pattern in ICE vehicles.}
    \label{fig:speed_spacing_comparison}
\end{figure}

The speed-spacing relationship during CF is illustrated in Figures~\ref{fig:ice_speed_spacing} and~\ref{fig:ev_speed_spacing} for ICE vehicles and EVs, respectively. ICE vehicle trajectories (VehA Min, VehA Max, VehB Min, VehB Max) exhibit annular, ring-like patterns, indicating weak speed-spacing correlation due to mechanical response limitations. Minimum ACC settings result in more compact distributions than maximum settings, suggesting smaller spacing preferences. In contrast, EV trajectories (Short 55, Medium 55, Long 55, Xlong 55) are elongated and follow a near-linear trend, particularly pronounced in Long 55 and Xlong 55. This behavior reflects the higher motor responsiveness and regenerative braking capabilities of EVs. Such dynamic characteristics present significant challenges for modeling EV behavior using traditional ICE-based CF models.

The analyses demonstrate that EV CF behavior, characterized by asymmetric acceleration, aggressive speed maintenance, larger spacing preferences, and stronger speed-spacing correlations, diverges notably from that of ICE vehicles. The rapid responsiveness of EVs, enabled by instantaneous torque and regenerative braking, presents challenges to traditional ICE-based CF models. These differences in vehicle dynamics highlight the need for AI-based approaches capable of distinguishing between ICE and EV behaviors, thereby supporting accurate mixed-traffic simulations across varying EV penetration levels. Collectively, these datasets contribute to advancing CF modeling in the context of vehicle electrification.

\subsubsection{Autocorrelation Analysis of Spacing}

The autocorrelation function (ACF) measures how spacing at time \( t \) correlates with spacing at a later lag \( k \), formally defined for a zero-mean series \( s_t \) with variance \( \sigma^2 \) as:
\begin{equation}
\rho(k) = \frac{\mathbb{E}\bigl[(s_t - \mu)(s_{t+k} - \mu)\bigr]}{\sigma^2}
\end{equation}
where \( \mu \) is the mean spacing. We compute \( \rho(k) \) up to \( k = 2000 \) samples to capture both short- and long-term dependencies, indicating the predictability of current spacing from past values. A value of \( \rho(k) \approx +1 \) indicates persistent large spacing, \( \rho(k) \approx -1 \) signifies an inverse relationship, and \( \rho(k) \approx 0 \) denotes weak or no correlation. A rapidly decaying \( \rho(k) \) reflects short memory, whereas a slowly decaying or oscillatory \( \rho(k) \) suggests persistent or cyclical patterns.

\begin{figure}[t!]
  \centering
  \begin{subfigure}[b]{0.23\linewidth}
    \centering
    \includegraphics[width=\linewidth]{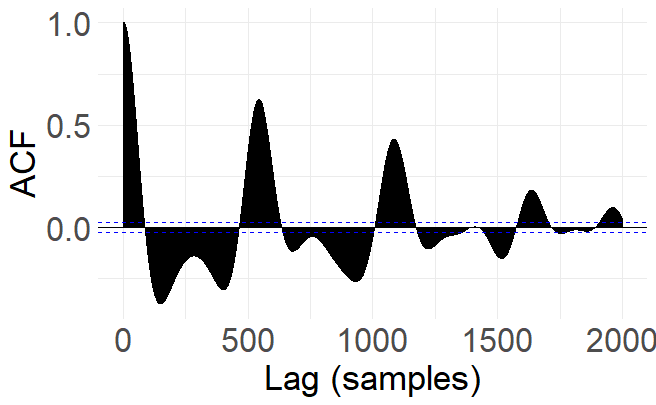}
    \caption{ICE VehA Min}
    \label{fig:acf_ice_A_min}
  \end{subfigure}
  \hfill
  \begin{subfigure}[b]{0.23\linewidth}
    \centering
    \includegraphics[width=\linewidth]{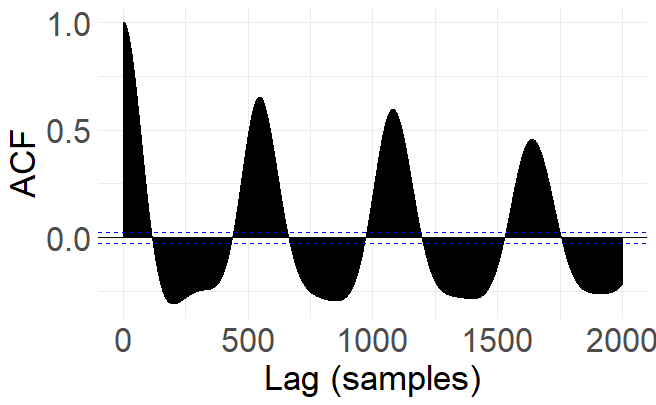}
    \caption{ICE VehA Max}
    \label{fig:acf_ice_A_max}
  \end{subfigure}
  \hfill
  \begin{subfigure}[b]{0.23\linewidth}
    \centering
    \includegraphics[width=\linewidth]{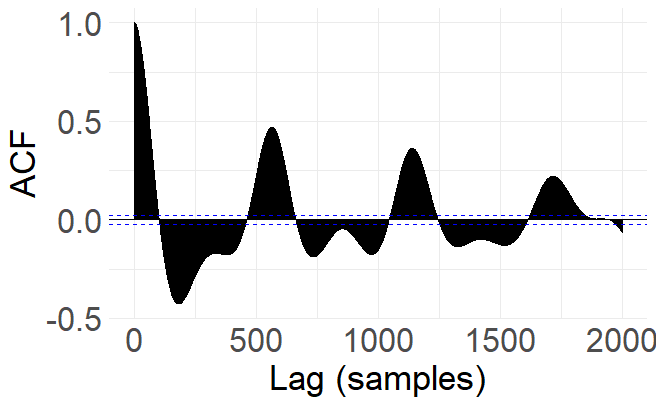}
    \caption{ICE VehB Min}
    \label{fig:acf_ice_B_min}
  \end{subfigure}
  \hfill
  \begin{subfigure}[b]{0.23\linewidth}
    \centering
    \includegraphics[width=\linewidth]{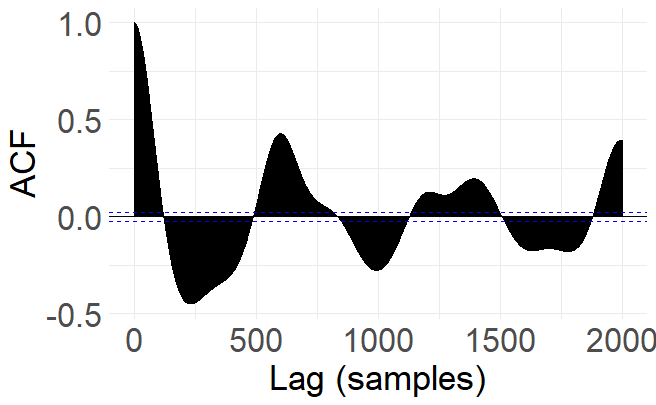}
    \caption{ICE VehB Max}
    \label{fig:acf_ice_B_max}
  \end{subfigure}

  \vspace{1em}

  \begin{subfigure}[b]{0.23\linewidth}
    \centering
    \includegraphics[width=\linewidth]{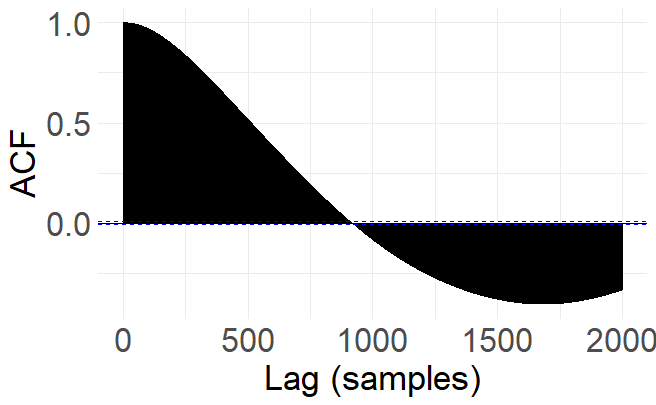}
    \caption{EV Short 55}
    \label{fig:acf_ev_short55}
  \end{subfigure}
  \hfill
  \begin{subfigure}[b]{0.23\linewidth}
    \centering
    \includegraphics[width=\linewidth]{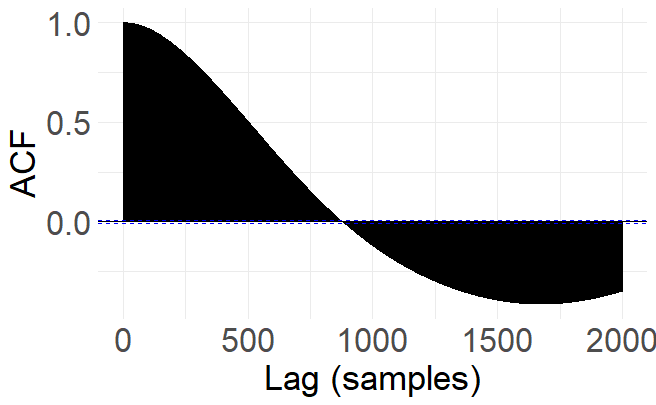}
    \caption{EV Medium 55}
    \label{fig:acf_ev_medium55}
  \end{subfigure}
  \hfill
  \begin{subfigure}[b]{0.23\linewidth}
    \centering
    \includegraphics[width=\linewidth]{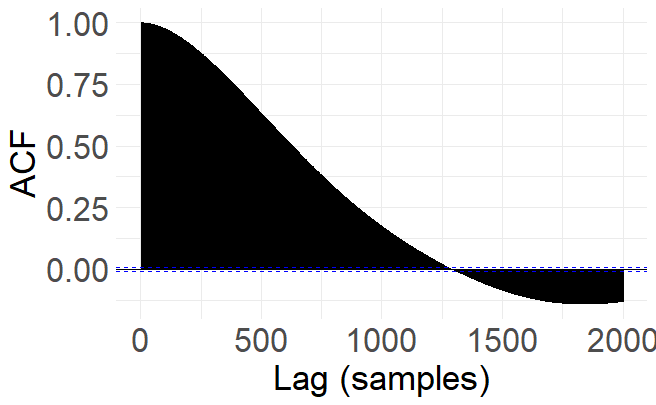}
    \caption{EV Long 55}
    \label{fig:acf_ev_long55}
  \end{subfigure}
  \hfill
  \begin{subfigure}[b]{0.23\linewidth}
    \centering
    \includegraphics[width=\linewidth]{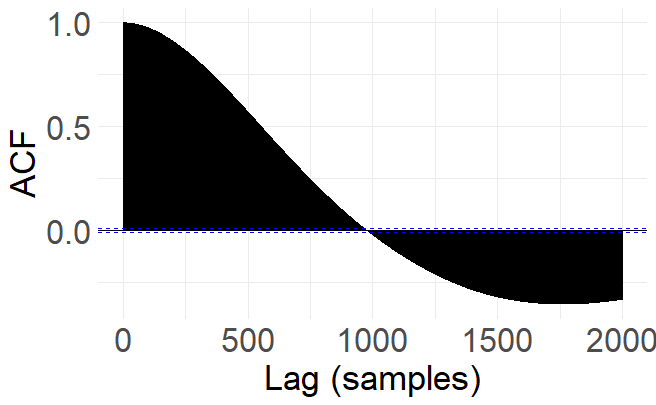}
    \caption{EV Xlong 55}
    \label{fig:acf_ev_xlong55}
  \end{subfigure}
  \caption{%
    Spacing‐ACF plots for all ICE (top row) and EV (bottom row) conditions. 
    (a) ICE VehA Min, (b) ICE VehA Max, (c) ICE VehB Min, (d) ICE VehB Max, 
    (e) EV Short 55, (f) EV Medium 55, (g) EV Long 55, and (h) EV Xlong 55.%
  }
  \label{fig:acf_all}
\end{figure}

Figure~\ref{fig:acf_all} presents spacing ACF plots for ICE vehicles (top row) and EVs (bottom row). For ICE vehicles under different ACC settings (Figures~\ref{fig:acf_ice_A_min}–\ref{fig:acf_ice_B_max}), ACF curves exhibit alternating positive and negative lobes with a period of approximately 100--450 samples. Specifically, the ACF of VehA Min dips to about -0.30 near lag 100, peaks at +0.60 around lag 500, and repeats, indicating recurring stop-and-go waves. This oscillatory behavior arises from aggressive throttle and friction braking cycles, where spacing grows during acceleration and shrinks during braking, cycling approximately every 100 seconds.

In contrast, EV spacing ACFs for Short 55, Medium 55, Long 55, and Xlong 55 (Figures~\ref{fig:acf_ev_short55}–\ref{fig:acf_ev_xlong55}) start at 1.0 (lag 0) and decline smoothly toward zero, crossing it around lag 1000, reaching a minimum of approximately -0.3, and slowly returning to zero. This monotonic decay suggests that once a target gap is achieved, spacing persists with minor adjustments, with a negative dip around lag 1500 indicating a slight opposite deviation after about 300 seconds. The absence of oscillatory lobes confirms the lack of strong periodic stop-and-go behavior in EVs.

These contrasting ACF shapes arise from differences in control logic and vehicle dynamics between the two vehicle types. ICE vehicles rely on discrete acceleration and deceleration cycles, leading to periodic spacing waves. In contrast, EVs employ rapid torque modulation and regenerative braking to enable continuous, fine-grained adjustments. This smoother, high-bandwidth control in EVs results in a more gradual ACF decay, reflecting their enhanced ability to maintain stable gaps relative to the cyclical patterns observed in ICE vehicles. These findings underscore the need for EV-specific CF models that can accurately capture such dynamic behaviors in mixed traffic simulations.

\section{Methodology}\label{sec:methodology}

Conventional CF models, developed for ICE vehicles, struggle to capture the complex microscopic traffic flow dynamics, such as nonlinear dynamics and rapid acceleration, that the increasing adoption of EVs introduces~\citep{rajamani2011vehicle, treiber2000congested, long2024traffic, mo2021physics}. These models rely on analytical formulations assuming uniform acceleration patterns, which are inadequate for EV-specific behaviors such as rapid acceleration and regenerative braking. Drawing from~\cite{long2024traffic}, which proposes a Physics-Enhanced Residual Learning (PERL) approach for improving traffic flow dynamics, incorporating residual corrections to enhance prediction, and~\cite{mo2021physics}, which introduces a Physics-Informed Neural Network (PINN) paradigm integrating mathematical formulations with neural networks to model car-following behavior, we identify that such limitations stem from the inability to adapt to EV powertrain dynamics. Such behaviors are critical for optimizing transportation systems in mixed traffic environments with EVs and ICE vehicles. To address this critical limitation, we propose an AI-based framework leveraging sequential neural networks and high-fidelity ACC trajectory data to model nonlinear interactions and phase transitions in real-world EV dynamics~\citep{hochreiter1997long, vaswani2017attention, long2024traffic, mo2021physics}. Building on~\cite{long2024traffic}'s predictive control strategy for enhancing traffic flow, with residual corrections integrated, and~\cite{mo2021physics}'s PINN approach for combining physics-based models with deep learning, our framework includes the proposed Phase-Aware AI (PAAI) model, complemented by the Baseline AI model for comparative analysis. The PAAI model addresses the asymmetric nature of EV dynamics, particularly the differences between rapid acceleration and regenerative braking, to enhance modeling accuracy and maintain physical consistency in mixed traffic scenarios. The following subsections outline the classical CF formulation, the proposed AI-based framework, and the design and training methodologies of the Baseline AI model and the PAAI model.

\subsection{Classical CF Framework}

Conventional CF models provide a framework for microscopic traffic flow analysis, modeling interactions between ICE vehicles through analytical relationships among key traffic variables~\citep{rajamani2011vehicle, treiber2000congested}. These models quantify acceleration as:
\begin{equation}\label{eq:acceleration}
    a(t) = f(\boldsymbol{\theta}, s(t), v(t), \Delta v(t))
\end{equation}
where \( a(t) \) is the acceleration (m/s\(^2\)) at time \( t \), \( f(\cdot) \) is the CF model function, \( \boldsymbol{\theta} \) represents model parameters, \( s(t) \) is the spacing (m), \( v(t) \) is the speed (m/s), and \( \Delta v(t) \) is the relative speed (m/s) to the lead vehicle. This formulation, based on established analytical approaches, assumes acceleration depends on spacing, speed, and relative speed~\citep{rajamani2011vehicle, treiber2000congested}. However, it struggles to capture EV-specific non-uniform dynamics, such as rapid acceleration and regenerative braking~\citep{li2020research}, necessitating data-driven methods for effective modeling of complex interactions. The following subsections present an AI-based framework to address these critical limitations.

\subsection{AI-Based CF Framework}

To address limitations of conventional CF models designed for ICE vehicles, which may fail to capture EV-specific nonlinear dynamics such as rapid acceleration and regenerative braking, we propose an AI-based framework leveraging sequential neural networks and high-fidelity ACC trajectory data~\citep{hochreiter1997long, vaswani2017attention}. The framework comprises two complementary models: the Baseline AI model, adopting a fully data-driven approach to learn nonlinear EV dynamics from trajectory data, and the PAAI model, integrating traditional CF models with AI corrections to refine them. The PAAI model explicitly models asymmetric dynamics, reflecting the distinct characteristics of rapid acceleration and regenerative braking in EVs, while ensuring physical consistency. Both models employ attention to enhance temporal feature selection and corrections---AI adjustments that improve the accuracy of traditional CF models in handling EV dynamics---to mitigate error propagation, ensuring robust performance in mixed traffic scenarios. The following subsections detail their design and training processes.

\subsubsection{Baseline AI model}

The Baseline AI model emphasizes flexibility in capturing EV-specific CF dynamics, such as rapid acceleration and regenerative braking, through a purely data-driven approach. It leverages a sequential neural network to implicitly learn acceleration and deceleration transitions from ACC trajectory data, ensuring adaptability across diverse traffic conditions without reliance on predefined physical models for ICE vehicles~\citep{hochreiter1997long}. The sequential neural network excels in processing time-series data and capturing complex temporal patterns, offering a robust foundation for modeling EV dynamics.

\paragraph{Model Design}
The model employs a sequential neural network with attention to prioritize salient temporal features for EV dynamics:
\begin{equation}\label{eq:baseline_acc}
    \hat{a}(t) = \mathrm{NN}(\mathbf{x}_{t-s+1:t}; \boldsymbol{\Theta})
\end{equation}
where \( \hat{a}(t) \) is the predicted acceleration (m/s\(^2\)) at time \( t \), \( \mathrm{NN}(\cdot) \) is the neural network function, \( \mathbf{x}_{t-s+1:t} \) is the input sequence over \( s \) time steps including traffic variables (e.g., speed, spacing, relative speed, safety margin), and \( \boldsymbol{\Theta} \) denotes learnable parameters. Attention computes weights as:
\begin{equation}\label{eq:baseline_attention}
    \mathbf{a}_t = \mathrm{softmax}\left( \mathbf{W}_a \cdot \mathrm{ReLU}\left( \mathbf{W}_b \cdot \mathbf{h}_t + \mathbf{b}_b \right) + \mathbf{b}_a \right)
\end{equation}
where \( \mathbf{a}_t \) is the attention weight vector at time \( t \), focusing on key temporal features such as acceleration differences, spacing variations, and speed changes in EV dynamics, \( \mathbf{W}_a \) and \( \mathbf{W}_b \) are weight matrices, \( \mathbf{h}_t \) is the sequential hidden state, \( \mathbf{b}_b \) and \( \mathbf{b}_a \) are bias vectors, and \( \mathrm{ReLU}(\cdot) \) is the rectified linear unit activation~\citep{vaswani2017attention}. Attended features are aggregated as:
\begin{equation}\label{eq:baseline_attended}
    \mathbf{h}_{\text{att}} = \sum_{i=t-s+1}^t \mathbf{a}_i \cdot \mathbf{h}_i
\end{equation}
where \( \mathbf{h}_{\text{att}} \) is the weighted sum of hidden states. Acceleration outputs are constrained to ensure physical realism, enhancing robustness for EV dynamics.

\paragraph{Training Methodology}
Training employs a continuous state framework with smoothed ACC data, using a loss function to balance acceleration accuracy, spacing accuracy, and stability for EV dynamics:
\begin{equation}\label{eq:loss_combined}
    \mathcal{L}(\boldsymbol{\Theta}) = \alpha \mathcal{L}_{\text{acc}} + \beta \mathcal{L}_{\text{safe}} + \gamma \mathcal{L}_{\text{reg}}
\end{equation}
where \( \mathcal{L}_{\text{acc}} \) measures acceleration accuracy using an error-based loss (e.g., smooth L1 or mean squared error (MSE), detailed in Section~\ref{sec:simulations}), \( \mathcal{L}_{\text{safe}} \) enforces physical constraints like safe spacing for EVs and ICE vehicles, \( \mathcal{L}_{\text{reg}} \) promotes stability via regularization, and \( \alpha, \beta, \gamma \) are weighting coefficients. An ensemble approach enhances robustness across diverse traffic conditions~\citep{breiman1996bagging}.

\subsubsection{PAAI model}

The PAAI model integrates traditional CF frameworks for ICE vehicles with AI corrections to capture EV-specific asymmetric dynamics, such as rapid acceleration and regenerative braking, while ensuring physical consistency. A key innovation lies in the phase-aware add-speed recognition, which dynamically identifies and distinguishes acceleration and deceleration phases using adaptive phase weights, addressing the distinct characteristics of EV motion. Another significant contribution is the residual correction, where AI refines traditional CF model predictions by correcting residual errors, enhancing adaptability to EV dynamics. These asymmetric dynamics arise from the distinct phases of acceleration and deceleration, with rapid acceleration contrasting regenerative braking. Its design combines the interpretability of physical models with the adaptability of data-driven methods, explicitly addressing phase transitions to enhance predictive accuracy in mixed traffic environments.

\paragraph{Model Design}
This hybrid approach predicts acceleration as:
\begin{equation}\label{eq:phase_aware_acc}
    \hat{a}(t) = a_{\text{base}}(t) + \mathrm{NN}(\mathbf{x}_{t-s+1:t}, w_{\text{phase}}; \boldsymbol{\Theta})
\end{equation}
where \( \hat{a}(t) \) is the predicted acceleration (m/s\(^2\)) at time \( t \), \( a_{\text{base}}(t) \) is the baseline acceleration (m/s\(^2\)) derived from a traditional CF model, e.g., the intelligent driver model (IDM)~\citep{treiber2000congested} or the optimal velocity with relative velocity (OVRV) model~\citep{milanes2013cooperative}, providing a physics-based foundation, \( w_{\text{phase}} \) is a phase weight distinguishing acceleration, deceleration, and phase transitions, and other terms are as defined in Equation~\eqref{eq:baseline_acc}. Phase transitions are identified as:
\begin{equation}\label{eq:phase_threshold}
    w_{\text{phase}} = g(\Delta v(t), m_s(t); \boldsymbol{\phi})
\end{equation}
where \( g(\cdot) \) determines the phase weight based on relative speed \( \Delta v(t) \) (m/s), safety margin \( m_s(t) \), and threshold parameters \( \boldsymbol{\phi} \). The computation of \( w_{\text{phase}} \) involves processing historical data of \( \Delta v(t) \) and \( m_s(t) \) through a normalization step using a sigmoid function, followed by a weighted sum with \( \boldsymbol{\phi} \) to classify the current traffic phase, resulting in \( w_{\text{phase}} \in [0, 1] \). Here, \( w_{\text{phase}} \) serves as the weight coefficient for the acceleration correction \( a_{\text{acc}} \), while \( 1 - w_{\text{phase}} \) is the weight coefficient for the deceleration correction \( a_{\text{dec}} \), used to combine \( a_{\text{acc}} \) and \( a_{\text{dec}} \) into the final neural network prediction \( a_{\text{nn}} \), as shown in Equation~\eqref{eq:phase_correction}. A value of \( w_{\text{phase}} = 1 \) indicates a pure acceleration phase, \( w_{\text{phase}} = 0 \) indicates a pure deceleration phase, and intermediate values represent transitional states with mixed contributions. The neural network incorporates baseline predictions in \( \mathbf{x}_{t-s+1:t} \), using attention from Equations~\eqref{eq:baseline_attention} and \eqref{eq:baseline_attended}. Separate attention for acceleration computes correction:
\begin{equation}\label{eq:phase_acc}
    a_{\text{acc}} = \mathbf{W}_{\text{acc}} \cdot \mathrm{ReLU}\left( \mathbf{W}_m \cdot \mathbf{h}_{\text{att,acc}} + \mathbf{b}_m \right) + \mathbf{b}_{\text{acc}}
\end{equation}
where \( a_{\text{acc}} \) represents the acceleration correction (m/s\(^2\)), computed by applying a weight matrix \( \mathbf{W}_{\text{acc}} \) to the ReLU-activated output of \( \mathbf{W}_m \cdot \mathbf{h}_{\text{att,acc}} + \mathbf{b}_m \), followed by a bias \( \mathbf{b}_{\text{acc}} \) to refine the prediction. The attention mechanism is implemented by taking \( \mathbf{h}_t \) from previous time steps, computing \( \mathbf{h}_{\text{att,acc}} \) through weighted sums based on \( \mathbf{a}_t \), and producing \( \mathbf{h}_{\text{att,acc}} \) for acceleration-specific correction. This process introduces nonlinearity via the ReLU activation, enabling the model to capture complex acceleration patterns, while the weight matrix and bias adjust the focus on phase-specific features to improve accuracy in modeling EV dynamics. Similarly, the deceleration correction (m/s\(^2\)) is given by:
\begin{equation}\label{eq:phase_dec}
    a_{\text{dec}} = \mathbf{W}_{\text{dec}} \cdot \mathrm{ReLU}\left( \mathbf{W}_n \cdot \mathbf{h}_{\text{att,dec}} + \mathbf{b}_n \right) + \mathbf{b}_{\text{dec}}
\end{equation}
computed similarly with \( \mathbf{W}_{\text{dec}} \), \( \mathbf{W}_n \), \( \mathbf{h}_{\text{att,dec}} \), \( \mathbf{b}_n \), and \( \mathbf{b}_{\text{dec}} \) to adjust for deceleration-specific dynamics. The ReLU activation here allows the model to learn nonlinear deceleration behaviors, and the learnable parameters fine-tune the model to reflect the asymmetric nature of EV motion, contributing to enhanced predictive performance. Here, \( \mathbf{h}_{\text{att,acc}}, \mathbf{h}_{\text{att,dec}} \) are attention-weighted hidden states, and \( \mathbf{W}_{\text{acc}}, \mathbf{W}_m, \mathbf{b}_m, \mathbf{b}_{\text{acc}}, \mathbf{W}_{\text{dec}}, \mathbf{W}_n, \mathbf{b}_n, \mathbf{b}_{\text{dec}} \) are learnable parameters. The final correction is:

\begin{equation}\label{eq:phase_correction}
    a_{\text{nn}} = w_{\text{phase}} a_{\text{acc}} + (1 - w_{\text{phase}}) a_{\text{dec}}
\end{equation}
where \( a_{\text{nn}} \) is the neural network correction (m/s\(^2\)), and \( w_{\text{phase}} \in [0, 1] \) is the phase weight, determined by relative speed \( \Delta v(t) \) and safety margin \( m_s(t) \) as defined in Equation \eqref{eq:phase_threshold}, balancing contributions from acceleration (\( a_{\text{acc}} \)) and deceleration (\( a_{\text{dec}} \)) to capture the asymmetric dynamics of EVs. This PAAI approach explicitly models asymmetric EV dynamics, enhancing adaptability and physical consistency. The step-by-step procedure of the algorithm is outlined in \textbf{Algorithm~1}.

\begin{table}[t!]
    \centering
    \caption*{\textbf{Algorithm~1}: PAAI Model}
    \label{tab:phase_aware_algorithm}
    \begin{tabular}{clr}
        \toprule
        \textbf{Input:} & \multicolumn{2}{l}{\(\mathbf{x}_{t-s+1:t}, a_{\text{base}}(t), \Delta v(t), m_s(t), \boldsymbol{\phi}\)} \\
        \textbf{Output:} & \multicolumn{2}{l}{\(\hat{a}(t)\)} \\
        \midrule
        1 & \(w_{\text{phase}} = g(\Delta v(t), m_s(t); \boldsymbol{\phi})\) & Compute phase weight \\
        2 & \(\sigma(w_{\text{phase}}) = \frac{1}{1 + e^{-w_{\text{phase}}}}\), \(w_{\text{phase}} \in [0, 1]\) & Normalize with sigmoid \\
        3 & \(\mathbf{h}_t = \text{LSTM}(\mathbf{x}_{t-s+1:t}; \boldsymbol{\Theta})\) & Process sequence with LSTM \\
        4 & \(\mathbf{a}_{\text{acc}} = \text{softmax}(\mathbf{W}_{\text{acc}} \cdot \text{ReLU}(\mathbf{W}_m \cdot \mathbf{h}_t + \mathbf{b}_m) + \mathbf{b}_{\text{acc}})\) & Acceleration attention weights \\
        5 & \(\mathbf{a}_{\text{dec}} = \text{softmax}(\mathbf{W}_{\text{dec}} \cdot \text{ReLU}(\mathbf{W}_n \cdot \mathbf{h}_t + \mathbf{b}_n) + \mathbf{b}_{\text{dec}})\) & Deceleration attention weights \\
        6 & \(\mathbf{h}_{\text{att,acc}} = \sum_{i=t-s+1}^{t} \mathbf{a}_{\text{acc},i} \cdot \mathbf{h}_i\) & Acceleration attention states \\
        7 & \(\mathbf{h}_{\text{att,dec}} = \sum_{i=t-s+1}^{t} \mathbf{a}_{\text{dec},i} \cdot \mathbf{h}_i\) & Deceleration attention states \\
        8 & \(a_{\text{acc}} = \mathbf{W}_{\text{acc}} \cdot \text{ReLU}(\mathbf{W}_m \cdot \mathbf{h}_{\text{att,acc}} + \mathbf{b}_m) + \mathbf{b}_{\text{acc}}\) & Acceleration correction \\
        9 & \(a_{\text{dec}} = \mathbf{W}_{\text{dec}} \cdot \text{ReLU}(\mathbf{W}_n \cdot \mathbf{h}_{\text{att,dec}} + \mathbf{b}_n) + \mathbf{b}_{\text{dec}}\) & Deceleration correction \\
        10 & \(a_{\text{nn}} = w_{\text{phase}} \cdot a_{\text{acc}} + (1 - w_{\text{phase}}) \cdot a_{\text{dec}}\) & Combine corrections \\
        11 & \(\hat{a}(t) = a_{\text{base}}(t) + a_{\text{nn}}\) & Compute final acceleration \\
        12 & \(\mathcal{L}(\boldsymbol{\Theta}) = \alpha \mathcal{L}_{\text{acc}} + \beta \mathcal{L}_{\text{safe}} + \gamma \mathcal{L}_{\text{reg}}\) & Compute loss function \\
        13 & \(\nabla_{\boldsymbol{\Theta}} \mathcal{L}(\boldsymbol{\Theta})\) & Compute gradients \\
        14 & \(\boldsymbol{\Theta} \gets \text{AdamW}(\boldsymbol{\Theta}, \nabla_{\boldsymbol{\Theta}} \mathcal{L}(\boldsymbol{\Theta}))\) & Update parameters \\
        15 & \textbf{return} \(\hat{a}(t)\) & Return predicted acceleration \\
        \bottomrule
    \end{tabular}
\end{table}

\paragraph{Training Methodology}
Training follows the continuous state framework and loss function of the Baseline AI model, defined in Equation~\eqref{eq:loss_combined}, with dynamic weights tuned to prioritize spacing accuracy for safe gap maintenance in mixed traffic environments involving EVs and ICE vehicles. Specific implementations of \( \mathcal{L}_{\text{acc}} \) are detailed in the subsequent Section~\ref{sec:simulations}. Unlike the Baseline AI model, which relies on a static loss structure, the PAAI model incorporates phase-aware adjustments to adapt to varying traffic phases, enhancing model flexibility as informed by the training data. This ensemble approach ensures robust performance.

\section{Experiment Results}\label{sec:simulations}

This section presents experimental results to evaluate the proposed AI-based CF models, building on the methodology developed in the preceding section. The goal is to validate their capability to capture EV-specific dynamics, such as rapid acceleration and regenerative braking, in mixed traffic environments containing both EVs and ICE vehicles. We assess both the Baseline AI model and the PAAI model against two traditional CF models for ICE vehicles, namely the OVRV model and the IDM, selected for their effectiveness and widespread use in the transportation community~\citep{rajamani2011vehicle,treiber2000congested}. Simulations are conducted in a ring road environment using ACC trajectory data to analyze both microscopic and macroscopic traffic metrics. Model performance is quantified using root mean square error (RMSE) metrics for acceleration, speed, and spacing, which are commonly used to evaluate CF behavior~\citep{punzo2016speed}.

\subsection{Experimental Setup and Baseline Models}\label{sec:setup}

Simulations used a real-world ACC dataset capturing EV trajectories in urban highway conditions~\citep{lapardhaja2023unlocking}. Two traditional CF models, the OVRV model and the IDM, were selected as baseline for comparison with AI-based models due to their established performance in modeling ICE vehicles~\citep{rajamani2011vehicle, treiber2000congested}. These models were compared to evaluate their effectiveness against AI-driven approaches. The OVRV model, chosen for its effectiveness in modeling ACC-equipped ICE vehicles, calculates acceleration as:
\begin{equation}\label{eq:ovrv}
a = k_1 (s - \eta - \tau v) + k_2 (v_l - v)
\end{equation}
where \( k_1 \) and \(k_2 \) are proportionality factors, \( \eta \) is the minimum safe distance (m), \( \tau \) is the desired time headway (s), \( s \) is the inter-vehicle spacing (m), \( v \) is the speed (m/s), and \( v_l \) is the lead vehicle speed (m/s). The IDM, selected for its adaptability to varying traffic conditions and validated performance in ICE-ACC systems, determines acceleration as:
\begin{equation}\label{eq:idm_acceleration}
a = \theta \left[ 1 - \left( \frac{v}{v_0} \right)^\delta - \left( \frac{\hat{s}(v, \Delta v)}{s} \right)^2 \right]
\end{equation}
\begin{equation}\label{eq:idm_spacing}
\hat{s}(v, \Delta v) = s_0 + T v + \frac{v \Delta v}{2 \sqrt{\theta \gamma}}
\end{equation}
where \( \theta \) is the maximum acceleration (m/s\(^2\)), \( v_0 \) is the free-flow speed (m/s), \( \delta \) is the acceleration exponent, \( s_0 \) is the minimum spacing (m), \( T \) is the time headway (s), \( \gamma \) is the comfortable deceleration (m/s\(^2\)), and \( \Delta v = v_l - v \) is the relative speed (m/s). Table~\ref{tab:model_params} lists optimized parameters, tuned via grid search to minimize spacing errors on the training set. These parameters are calibrated using real-world data, optimized to minimize the spacing RMSE, calculated as~\citep{punzo2016speed}:
\begin{equation}
\text{RMSE}_{\text{spacing}} = \sqrt{\frac{1}{N} \sum_{i=1}^N (s_{\text{pred}}(t_i) - s_{\text{actual}}(t_i))^2}
\end{equation}
where \( s_{\text{pred}}(t_i) \) and \( s_{\text{actual}}(t_i) \) are predicted and actual spacings (m), respectively, and \( N \) is the number of samples.
\begin{table}[t!]
    \captionsetup{justification=centering}
    \caption{Parameters and RMSE for the OVRV and IDM models.}
    \label{tab:model_params}
    \centering
    \small
    \setlength{\tabcolsep}{4pt}
    \begin{tabular}{@{}>{\centering\arraybackslash}m{3.0cm}>{\centering\arraybackslash}m{3.0cm}>{\centering\arraybackslash}m{3.0cm}@{}}
        \toprule
        \textbf{Model} & \textbf{Parameters} & \textbf{RMSE (m)} \\
        \midrule
        \multirow{4}{*}{OVRV} & \( k_1 = 0.0717 \) & \multirow{4}{*}{2.0648} \\
                              & \( k_2 = 0.6541 \) & \\
                              & \( \eta = 17.9107 \) & \\
                              & \( \tau = 0.5452 \) & \\
        \midrule
        \multirow{6}{*}{IDM} & \( \theta = 1.6932 \) & \multirow{6}{*}{2.5077} \\
                              & \( \gamma = 10.0000 \) & \\
                              & \( \delta = 5.0000 \) & \\
                              & \( s_0 = 6.0000 \) & \\
                              & \( T = 1.0325 \) & \\
                              & \( v_0 = 40.0000 \) & \\
        \bottomrule
    \end{tabular}
\end{table}

\subsection{Simulations}

This subsection outlines the simulation and training procedures used to validate the ability of AI-based CF models to capture EV-specific dynamics, e.g., rapid acceleration and regenerative braking, in mixed traffic environments. Simulations were conducted using preprocessed ACC data, which were split into 60\% training, 25\% validation, and 15\% testing sets.

\paragraph{Model Training}
The comprehensive loss function defined in Equation~\eqref{eq:loss_combined} is used for training to optimize EV-specific dynamics in mixed EV and ICE traffic environments:
\begin{equation}
    \mathcal{L}(\boldsymbol{\Theta}) = \alpha \mathcal{L}_{\text{acc}} + \beta \mathcal{L}_{\text{safe}} + \gamma \mathcal{L}_{\text{reg}}
\end{equation}
with the weights \(\alpha, \beta, and \gamma\) dynamically adjusted to prioritize spacing accuracy (e.g., \(\alpha = 0.5, \beta = 0.4, and \gamma = 0.1\) for Baseline AI; varying for the PAAI model based on epochs). For the Baseline AI model, phase transitions are captured implicitly through the temporal patterns of the Long Short-Term Memory (LSTM) without a phase weight~\citep{hochreiter1997long}. For PAAI models, a phase weight \( w_{\text{phase}} \) explicitly distinguishes acceleration, deceleration, and neutral phases:

\begin{equation}
    w_{\text{phase}} = \begin{cases} 
        0.8 & \text{if } \Delta v(t) > 0.1, \\
        0.2 & \text{if } \Delta v(t) < -0.1, \\
        0.5 & \text{otherwise}
    \end{cases}
\end{equation}
where \(\Delta v(t)\) is the relative speed, primarily used to determine the phase of the car-following behavior. The weights, 0.8, 0.2, and 0.5, were chosen to reflect the asymmetric dynamics of EVs, as implemented in the PAAI model. When \(\Delta v(t) > 0.1\), the weight of 0.8 emphasizes acceleration phases to capture the rapid torque response of EVs. When \(\Delta v(t) < -0.1\), the weight of 0.2 prioritizes the deceleration phase to capture the smoother regenerative braking behavior of EVs. For neutral phases, a weight of 0.5 provides a balanced contribution, ensuring flexibility in mixed traffic scenarios. To satisfy safety constraints, an additional safety margin is applied, defined as:
\begin{equation}
    m_s(t) = \frac{s(t) - s_{\text{safe}}(t)}{s_{\text{safe}}(t)}
\end{equation}
where the safe spacing is given by:
\begin{equation}
    s_{\text{safe}}(t) = 1.5 v(t) + 2
\end{equation}
Specifically, \(m_s(t) > 2.0\) is required for the acceleration phase to ensure safe spacing, while \(m_s(t) < -1.0\) is required for the deceleration phase to account for close car-following conditions.

The choice of acceleration loss \(\mathcal{L}_{\text{acc}}\) is tailored to each model's characteristics to effectively capture EV nonlinear dynamics, such as rapid torque response and regenerative braking. Baseline AI, being a general AI model, and OVRV PAAI, based on linear dynamics, use smooth L1 loss to balance precision for small errors and robustness against ACC data noise, suitable for their flexibility in modeling complex, heterogeneous EV behaviors~\citep{girshick2015fast}. Conversely, IDM PAAI employs MSE to leverage its smoother acceleration profiles, as the quadratic penalty of MSE enhances optimization of continuous trajectories, better fitting EV dynamics. For Baseline AI and OVRV PAAI, the loss is:
\begin{equation}
    \mathcal{L}_{\text{acc}}(x, y) = \begin{cases} 
        0.5 (x - y)^2 & \text{if } |x - y| < 1, \\
        |x - y| - 0.5 & \text{otherwise}
    \end{cases}
\end{equation}
where \( x \) and \( y \) are predicted and actual acceleration values (m/s\(^2\)), respectively, balancing precision for small errors and robustness against ACC data noise~\citep{girshick2015fast}. The IDM PAAI model uses:
\begin{equation}\label{eq:idm_acc_loss}
    \mathcal{L}_{\text{acc}}(x, y) = (x - y)^2
\end{equation}
to correct the base IDM model's smoother acceleration profiles, enabling better adaptation to the nonlinear dynamics of EVs, such as rapid torque response and regenerative braking. The safety loss \(\mathcal{L}_{\text{safe}}\) penalizes deviations from spacing:
\begin{equation}
    \mathcal{L}_{\text{safe}} = \frac{1}{N} \sum_{i=1}^N \max \left(0, s_{\text{safe}}(t_i) - s(t_i) \right)
\end{equation}
where \(s_{\text{safe}}(t_i) = 1.2 v(t_i) + 2\) for the Baseline AI and OVRV PAAI models, while \(s_{\text{safe}}(t_i) = 1.0 v(t_i)\) for the IDM PAAI models~\citep{treiber2000congested}. The regularization loss is defined as:
\begin{equation}
    \mathcal{L}_{\text{reg}} = \begin{cases}
        \text{Var}(\hat{a}(t)) + 10^{-5} \sum_{i} \|\boldsymbol{\Theta}_i\|^2, ~~~ \text{for Baseline AI and OVRV PAAI models}, \\
        \frac{1}{N} \sum_{i=1}^N |\hat{a}(t_i) - a(t_i)| + 10^{-5} \sum_{i} \|\boldsymbol{\Theta}_i\|^2, ~~~ \text{for IDM PAAI model}
    \end{cases}
\end{equation}
The regularization loss \(\mathcal{L}_{\text{reg}}\) differs to align with each model's training objectives. For the Baseline AI and OVRV PAAI models, designed directly for EV nonlinear dynamics, the use of acceleration variance promotes smoother predictions to mitigate the impact of noise and rapid changes in EV data, e.g., rapid torque response and regenerative braking. In contrast, the IDM PAAI model, despite simulating EVs, uses the mean absolute error (MAE) in \(\mathcal{L}_{\text{reg}}\), as implemented in the loss function of the PAAI model, to correct the base IDM model's tendency to produce smoother acceleration profiles, originally designed for ICE vehicles. The MAE emphasizes direct error between predicted and actual acceleration, enhancing adaptation to EV's high dynamic fluctuations. An ensemble of five models per approach was trained using AdamW optimization with a learning rate of 0.001, applying early stopping based on \( \mathcal{L}(\boldsymbol{\Theta}) \) for robust convergence~\citep{kingma2014adam}. Vehicle states were updated as:
\begin{equation}
    v(t + \Delta t) = v(t) + \hat{a}(t) \Delta t
\end{equation}
\begin{equation}
    s(t + \Delta t) = s(t) + (v_l(t) - v(t)) \Delta t
\end{equation}
where \(\Delta t = 0.1\) s, \( v(t) \) is the speed (m/s), and \( v_l(t) \) is the lead vehicle speed (m/s). 

\subsection{Results}

\begin{figure}[t!]
    \centering
    \begin{subfigure}{0.32\textwidth}
        \centering
        \includegraphics[width=\textwidth]{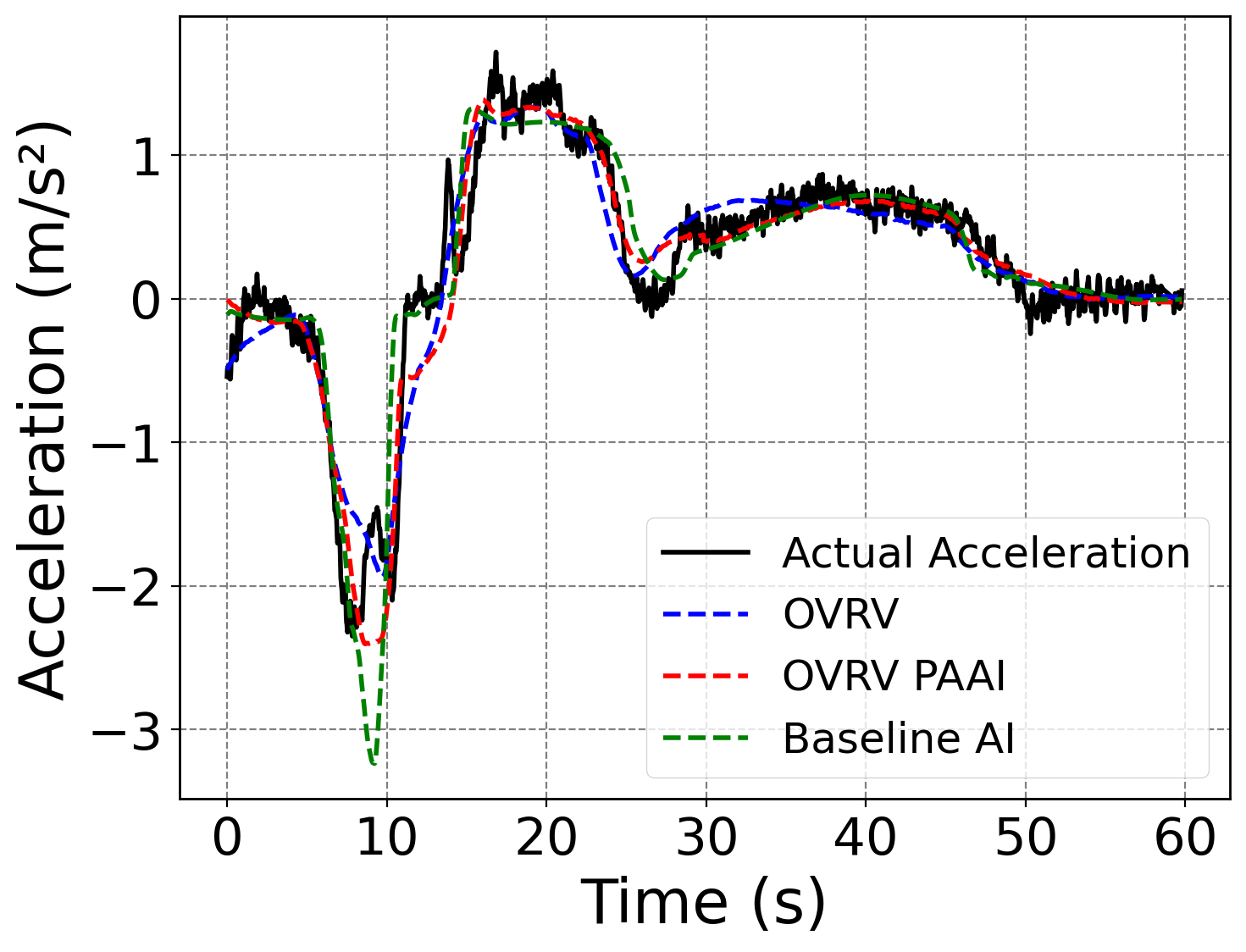}
        \caption{Acceleration Comparison (OVRV)}
        \label{fig:ovrv_acc}
    \end{subfigure}
    \hfill
    \begin{subfigure}{0.32\textwidth}
        \centering
        \includegraphics[width=\textwidth]{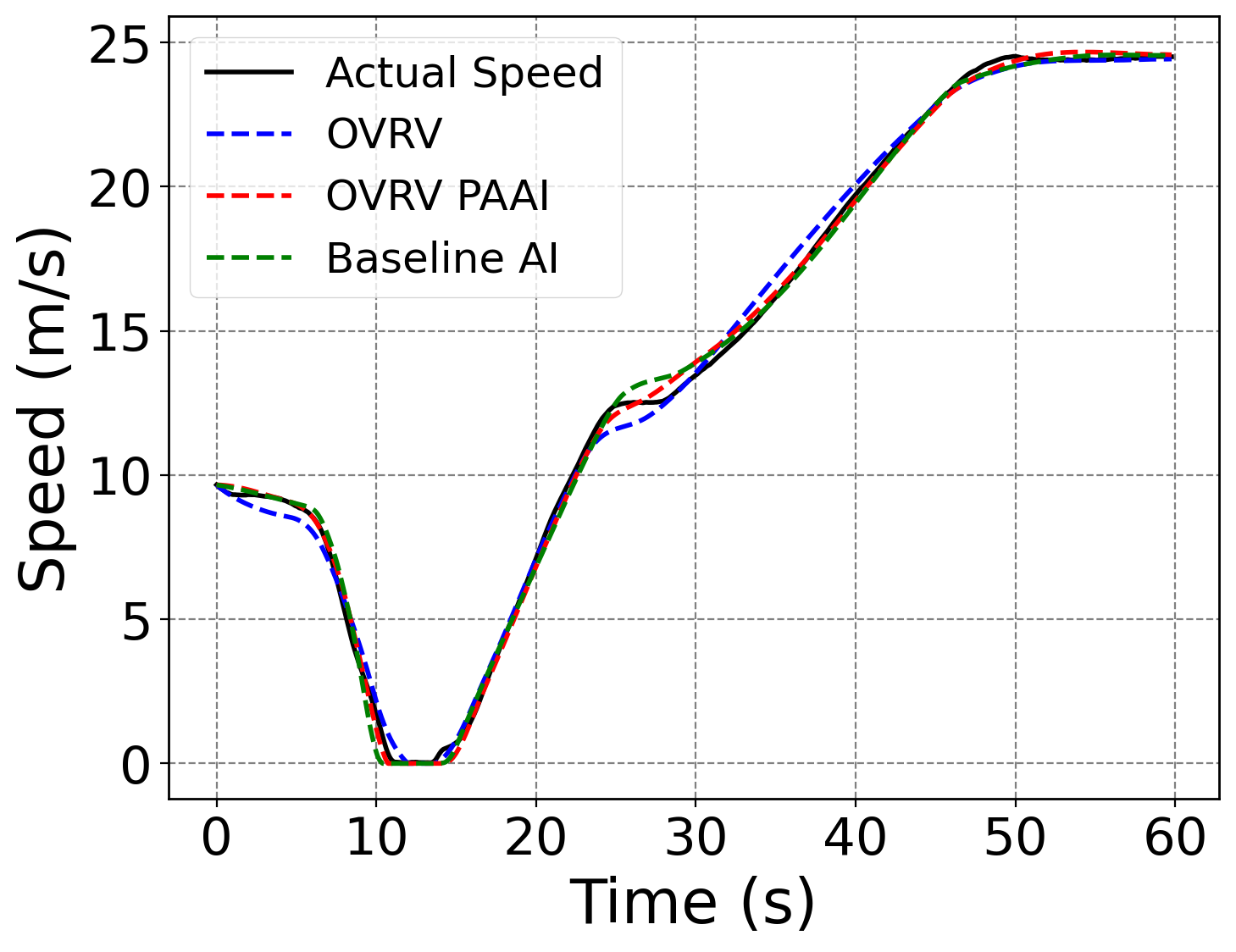}
        \caption{Speed Comparison (OVRV)}
        \label{fig:ovrv_speed}
    \end{subfigure}
    \hfill
    \begin{subfigure}{0.32\textwidth}
        \centering
        \includegraphics[width=\textwidth]{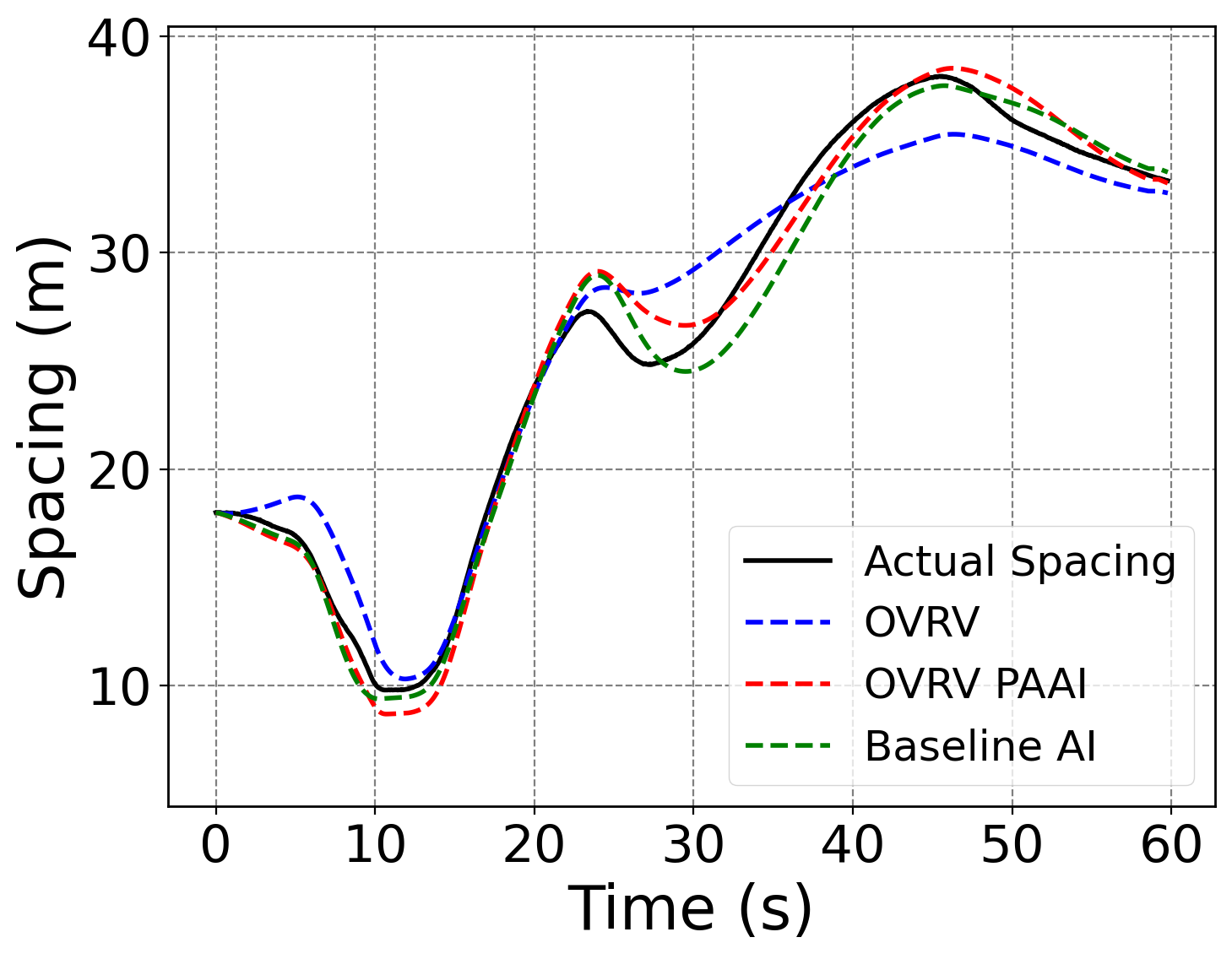}
        \caption{Spacing Comparison (OVRV)}
        \label{fig:ovrv_spacing}
    \end{subfigure}
    \vspace{0.5cm}
    \begin{subfigure}{0.32\textwidth}
        \centering
        \includegraphics[width=\textwidth]{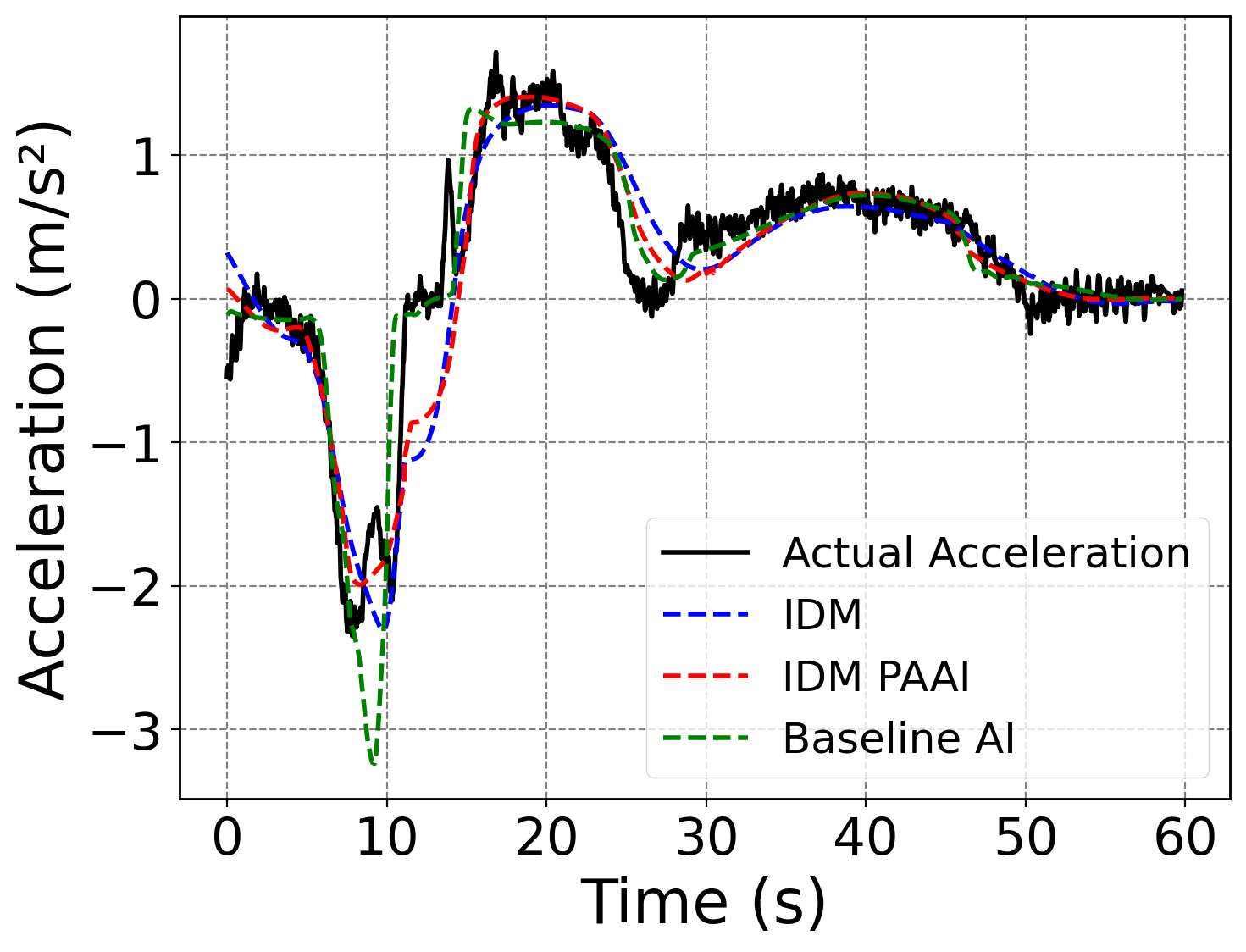}
        \caption{Acceleration Comparison (IDM)}
        \label{fig:idm_acc}
    \end{subfigure}
    \hfill
    \begin{subfigure}{0.32\textwidth}
        \centering
        \includegraphics[width=\textwidth]{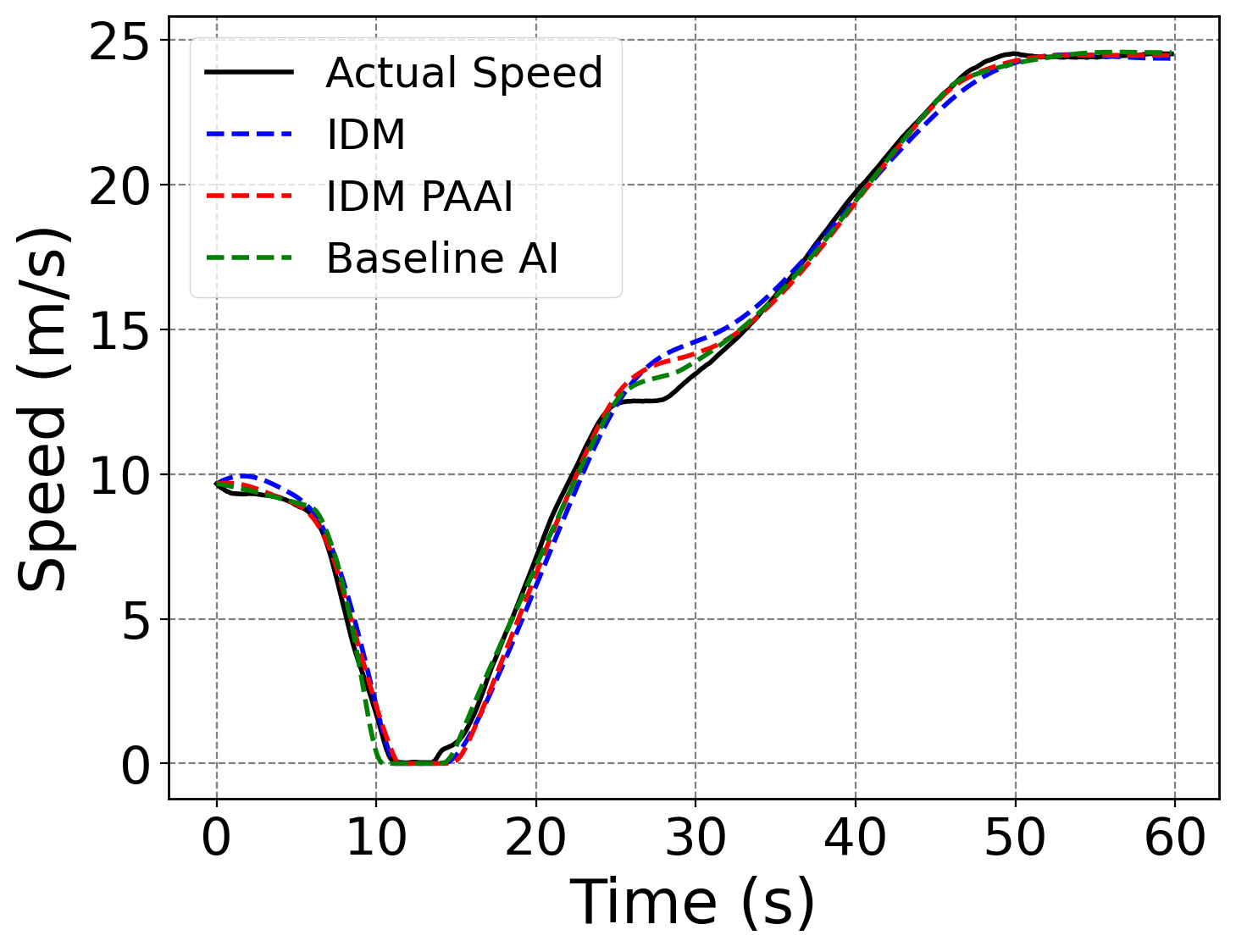}
        \caption{Speed Comparison (IDM)}
        \label{fig:idm_speed}
    \end{subfigure}
    \hfill
    \begin{subfigure}{0.32\textwidth}
        \centering
        \includegraphics[width=\textwidth]{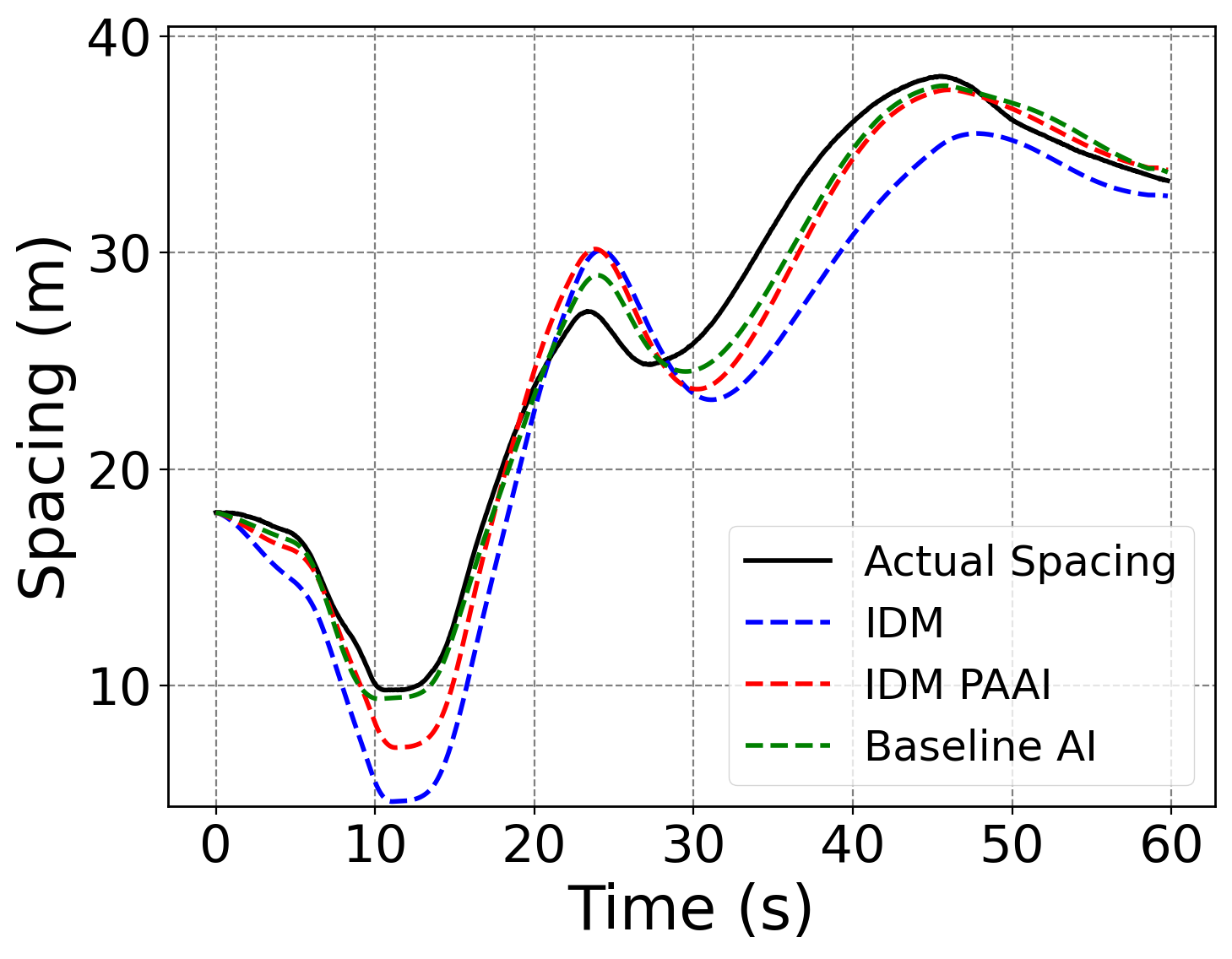}
        \caption{Spacing Comparison (IDM)}
        \label{fig:idm_spacing}
    \end{subfigure}
    \caption{Comparison of CF Model Predictions}
    \label{fig:comparison}
\end{figure}

Figure~\ref{fig:comparison} shows that acceleration prediction curves exhibit inherent high-frequency fluctuations, presenting a significant challenge in car-following modeling. The OVRV PAAI model demonstrates superior tracking capability, with its prediction curve closely aligning with the high-frequency features of actual acceleration and accurately reproducing instantaneous changes. This advantage is particularly evident during the sharp deceleration phase (5--10 seconds), where actual acceleration reaches approximately -2.5~m/s\(^2\). While all models capture the primary deceleration trend, the OVRV PAAI model excels in preserving detailed fluctuations that traditional CF models tend to smooth out. The acceleration initiation phase (10--15 seconds) further highlights this strength, as acceleration transitions rapidly from negative to positive values, peaking at approximately 1.5~m/s\(^2\). AI-enhanced models, especially the OVRV PAAI model, demonstrate superior accuracy in both peak prediction and timing precision. In contrast, the traditional IDM and IDM PAAI models exhibit noticeable lag, producing smoother predictions that lack sensitivity to these rapid transitions. The subsequent fluctuation adjustment phase (20--40 seconds) features multiple positive-to-negative oscillations in actual acceleration, which the OVRV PAAI and Baseline AI models track effectively, while traditional models' excessive smoothing fails to capture these critical dynamics.

In addition, Figure~\ref{fig:comparison} demonstrates systematic differences in speed prediction capabilities across the evaluated models. The speed profile follows a characteristic ``decline-then-rise'' pattern, decreasing from an initial 10~m/s to near 0~m/s before accelerating to a stable 24~m/s. During the initial deceleration phase (0--10 seconds), all models demonstrate effective tracking of the speed decline, though disparities become increasingly evident as the trajectory approaches its minimum value. The critical minimum point (10--15 seconds) represents a pivotal moment where the AI-enhanced models, particularly the OVRV PAAI model, exhibit superior precision in predicting both the minimum value and the exact timing of the turning point. This enhanced accuracy at critical transitions underscores the models' improved capability to handle nonlinear dynamics compared to traditional approaches. The subsequent acceleration phase (15--40 seconds) is characterized by pronounced nonlinear changes that further distinguish model performance. The OVRV PAAI model excels during this phase, with its prediction curve nearly overlapping the actual speed profile. Traditional IDM models, however, display systematic bias, particularly in the later acceleration stages where predicted values consistently fall below actual values, potentially impacting traffic flow predictions. During the stabilization phase (40--60 seconds), all models maintain relatively stable predictions, but the AI-enhanced models demonstrate better adaptability and accuracy in handling subtle fluctuations that persist even during stable driving conditions.

Considering spacing prediction and its safety implications, Figure~\ref{fig:comparison} elucidates each model's capability in controlling safe distances, with spacing trends inversely correlating with speed patterns. This inverse relationship, where spacing decreases as speed drops and increases as speed rises, aligns with expected driving behavior and validates the physical realism of the modeling approach. The initial phase (0--10 seconds) shows all models effectively tracking the spacing decrease from approximately 18~m. However, the critical minimum spacing point (10--15 seconds) represents a potentially hazardous moment where actual spacing reaches about 10~m. At this critical juncture, AI-enhanced models, especially the OVRV PAAI model, achieve higher prediction accuracy in both the minimum value and its timing. Traditional IDM models show noticeable deviations in minimum spacing prediction, potentially overestimating or underestimating actual safe distances, which could have serious safety implications. The spacing recovery phase (15--40 seconds) demonstrates the models' ability to predict distance management as vehicles accelerate. The OVRV PAAI model excels in predicting both the growth trend and dynamic variations during this phase, while traditional models show less sensitivity to the fluctuations that characterize real spacing behavior. In the stabilization phase (40--60 seconds), spacing stabilizes at approximately 35~m, but the traditional IDM model exhibits systematic bias with persistent differences from actual values, potentially impacting traffic safety assessment.

\begin{table}[t!]
    \centering
    \caption{RMSE comparison of CF models.}
    \label{tab:rmse}
    \begin{tabular}{ccccccc}
        \toprule
        Metric       & OVRV    & OVRV PAAI & IDM     & IDM PAAI & Baseline AI \\
        \midrule
        Acceleration (m/s\(^2\)) & 0.2331  & 0.2328        & 0.3298  & 0.2805       & 0.3241      \\
        Speed (m/s)  & 0.3818  & 0.2503        & 0.5747  & 0.4118       & 0.3287      \\
        Spacing (m)  & 1.8340  & 1.0546        & 3.4308  & 1.8089       & 1.1434      \\
        \bottomrule
    \end{tabular}
\end{table}

To provide a quantitative perspective, Table~\ref{tab:rmse} offers a detailed assessment based on the RMSE metric. The RMSE is defined as:
\begin{equation}
    \text{RMSE}_{\text{metric}} = \sqrt{\frac{1}{N} \sum_{i=1}^N (\hat{y}(t_i) - y_{\text{actual}}(t_i))^2}
\end{equation}
where \(\hat{y}(t_i)\) and \(y_{\text{actual}}(t_i)\) represent predicted and actual values for acceleration (m/s\(^2\)), speed (m/s), or spacing (m), respectively. For acceleration, the OVRV model achieves a slight performance advantage over its hybrid counterpart with an RMSE of 0.2331 compared to 0.2328 for the OVRV PAAI model, a marginal 0.13\% improvement, suggesting the pure physics-based approach suffices for this metric. The IDM model, however, benefits significantly from AI enhancement, reducing its RMSE from 0.3298 to 0.2805 for the IDM PAAI model---a 15\% improvement---while the Baseline AI model's RMSE of 0.3241 remains competitive but lags slightly due to noise. Speed prediction highlights the most substantial gains from AI integration, with the OVRV PAAI model achieving an RMSE of 0.2503, a 34.4\% improvement over the OVRV model's 0.3818, and the IDM PAAI model improving by 28.3\% from 0.5747 to 0.4118. The Baseline AI model, with an RMSE of 0.3287, outperforms both pure models, reflecting the general effectiveness of AI enhancement. Spacing, a critical safety metric, shows the most dramatic improvements, with the OVRV PAAI model recording an RMSE of 1.0546---a 42.5\% reduction from the OVRV model's 1.8340---and the IDM PAAI model achieving a 47.3\% decrease from 3.4308 to 1.8089. The Baseline AI model's RMSE of 1.1434, improved by 37.6\% over OVRV and 66.7\% over IDM, lags behind the OVRV PAAI model by 8.5\%, emphasizing the pivotal role of physics-based corrections in mitigating drift.

The transformative impact of AI enhancement, as further analyzed from Table~\ref{tab:rmse}, is evident across all model types. The IDM model improves from an RMSE of 0.3298 to 0.2805 for acceleration, 0.5747 to 0.4118 for speed, and 3.4308 to 1.8089 for spacing with the IDM PAAI model, while the OVRV model advances from 0.2331 to 0.2328 for acceleration, 0.3818 to 0.2503 for speed, and 1.8340 to 1.0546 for spacing with the OVRV PAAI model. These gains are particularly pronounced in speed, with the IDM PAAI model improving by 28.3\% and the OVRV PAAI model by 34.4\%, as well as in spacing, where the IDM PAAI model sees a 47.3\% reduction and the OVRV PAAI model a 42.5\% decrease. This consistent performance uplift suggests that AI enhancements effectively adapt to the complexities of real driving behavior, excelling in managing nonlinear dynamics and random fluctuations that characterize human driving patterns. The varying degrees of improvement across different base models indicate that the theoretical framework of the base model, combined with the tailored AI enhancement strategy, critically shapes the final performance outcome. The OVRV PAAI model's outstanding performance reflects both the effectiveness of AI integration and the inherent suitability of the OVRV framework for modeling complex driving behaviors, suggesting that the choice of base model remains a key factor even with AI enhancement.

\section{Conclusions}\label{sec:conclusion}

This study introduces a novel AI-based framework for CF modeling, addressing the shortcomings of traditional models in capturing EV-specific behaviors, such as rapid torque response and regenerative braking, in mixed traffic with EVs and ICE vehicles. The primary contribution is the development of the PAAI model, which combines traditional CF models, like the OVRV and IDM, with phase-aware recognition to accurately distinguish EV acceleration and deceleration phases by analyzing speed differences and safety margins. The PAAI model incorporates residual correction to refine predictions, ensuring physical consistency while capturing EV-specific dynamics, such as quick torque response and regenerative braking effects. This approach significantly improves prediction accuracy over traditional models and the Baseline AI model, particularly in mixed traffic scenarios with diverse vehicle behaviors. Simulations using real-world ACC trajectory data across various traffic conditions validate the PAAI model's reliability, showing superior performance in modeling asymmetric EV acceleration and deceleration patterns. Furthermore, the PAAI framework is adaptable and can be extended to various traffic scenarios, such as urban roads and intersections, thereby laying a foundation for advanced intelligent transportation systems.

However, the framework's performance faces challenges due to dataset limitations. Sparse or noisy datasets reduce model accuracy, particularly in dynamic traffic conditions like peak hours or rapidly changing flows. The lack of diverse EV trajectory data, covering different driving styles, weather conditions, road types, or driver demographics, limits the model's ability to apply broadly. Inconsistent data labeling, such as errors in manual annotations, further affects training outcomes. The model's reliance on specific EV types, such as high-performance models, may also limit its performance for other vehicle classes. Additionally, the PAAI model's computational demands pose challenges for real-time use, though data issues remain the primary concern.

In future work, we aim to address the current dataset limitations by conducting extensive EV CF experiments across diverse driving scenarios, including highways, urban roads, night conditions, and adverse weather, using various EV types such as sedans and SUVs. These experiments will enable the collection of high-quality, diverse trajectory data to improve model robustness and generalizability. To foster broader collaboration and consistent model validation, we plan to publicly share these datasets in standardized formats. We also intend to develop data enhancement techniques, such as synthetic data generation, to address gaps in empirical observations. Additionally, we will explore simplified model architectures to reduce computational demands and integrate the framework with intelligent transportation systems, including vehicle-to-everything communication, to improve its applicability in complex environments such as urban intersections.

\section*{Acknowledgment}
This work is supported by the Taylor Geospatial Institute and the President's Research Fund at Saint Louis University.



\clearpage
\newpage

\bibliographystyle{elsarticle-harv}

\bibliography{reference}
\end{document}